\documentclass{article}

\usepackage[final]{gpsmdms_2022}

\usepackage[round]{natbib}

\usepackage[utf8]{inputenc} 
\usepackage[T1]{fontenc}    
\usepackage{hyperref}       
\usepackage{url}            
\usepackage{booktabs}       
\usepackage{amsfonts}       
\usepackage{nicefrac}       
\usepackage{microtype}      
\usepackage{xcolor}         

\usepackage[utf8]{inputenc} %
\usepackage[T1]{fontenc}    %
\usepackage{hyperref}       %
\usepackage{url}            %
\usepackage{booktabs}       %
\usepackage{amsfonts}       %
\usepackage{nicefrac}       %
\usepackage{microtype}      %
\usepackage{xcolor}         %
\usepackage{adjustbox}
\usepackage{enumitem}
\usepackage{multirow}
\usepackage{diagbox}
\usepackage{amsmath}

\usepackage{caption}

\usepackage{tikz}
\usetikzlibrary{fit,positioning}

\usepackage{dsfont}

\usepackage{graphicx} %
\usepackage{amssymb}
\usepackage{times}
\usepackage{epsfig}
\usepackage{graphicx}
\usepackage{color}
\usepackage{bbm}
\usepackage{multicol}
\usepackage{natbib}
\usepackage{wrapfig}

\providecommand{\hideit}[1]{}

\usepackage{rotating}

\let\proof\relax
\let\endproof\relax
\usepackage{amsthm}
\usepackage{amsopn,amssymb,thmtools,thm-restate}

\newtheorem{thm}{Theorem}%
\newtheorem{lem}[thm]{Lemma}

\newtheorem*{remark}{Remark}

\usepackage{bm} %
\usepackage{algorithm}%
\usepackage{algpseudocode}%

\newcommand{\field}[1]{\mathbb{#1}}
\newcommand{\R}{\field{R}} %

\newcommand{\ProbOpr}[1]{\mathbb{#1}}

\newcommand{\expect}[2]{%
\ifthenelse{\equal{#2}{}}{\ProbOpr{E}_{#1}}
{\ifthenelse{\equal{#1}{}}{\ProbOpr{E}\left[#2\right]}{\ProbOpr{E}_{#1}\left[#2\right]}}} %
\newcommand{\var}[2]{%
\ifthenelse{\equal{#2}{}}{\ProbOpr{VAR}_{#1}}
{\ifthenelse{\equal{#1}{}}{\ProbOpr{VAR}\left[#2\right]}{\ProbOpr{VAR}_{#1}\left[#2\right]}}} %

\DeclareMathOperator{\argmax}{arg\,max}
\DeclareMathOperator{\argmin}{arg\,min}

\def\N{{\mathcal{N}}}
\def\GP{\mathcal{GP}}

\newcount\Comments  %
\usepackage{color}
\definecolor{darkred}{rgb}{0.5,0,0}
\definecolor{orange}{rgb}{1.0,0.64,0}
\definecolor{darkgreen}{rgb}{0,0.5,0}
\definecolor{darkblue}{rgb}{0,0,0.7}
\definecolor{purple}{rgb}{.6, 0,.6}
\newcommand{\kibitz}[2]{\ifnum\Comments=1\textcolor{#1}{#2}\fi}

\hypersetup{
  colorlinks = true,
  citecolor  = darkblue,
  linkcolor  = darkblue,
  citecolor  = darkblue,
  filecolor  = darkblue,
  urlcolor   = darkblue,
}

\usepackage{subcaption}
\usepackage{breqn}

\title{HyperBO+: Pre-training a universal prior for Bayesian optimization with hierarchical Gaussian processes}
\author{%
  Zhou Fan \\
  Harvard University\\
  \texttt{zfan@g.harvard.edu} \\
  \And
  Xinran Han \\
  Harvard University \\
  \texttt{xinranhan@g.harvard.edu} \\
  \AND
  Zi Wang \\
  Google Research, Brain Team \\
  \texttt{wangzi@google.com} \\
}

\begin{document}

\maketitle

\begin{abstract}

Bayesian optimization (BO), while proved highly effective for many black-box function optimization tasks, requires practitioners to carefully select priors that well model their functions of interest. Rather than specifying by hand, researchers have investigated transfer learning based methods to automatically learn the priors, e.g. multi-task BO~\citep{swersky2013multi}, few-shot BO~\citep{wistuba2021few} and HyperBO~\citep{wang2022pre}. However, those prior learning methods typically assume that the input domains are the same for all tasks, weakening their ability to use observations on functions with different domains or generalize the learned priors to BO on different search spaces. In this work, we present HyperBO+: a pre-training approach for hierarchical Gaussian processes that enables the same prior to work universally for Bayesian optimization on functions with different domains. We propose a two-step pre-training method and analyze its appealing asymptotic properties and benefits to BO both theoretically and empirically. On real-world hyperparameter tuning tasks that involve multiple search spaces, we demonstrate that HyperBO+ is able to generalize to unseen search spaces and achieves lower regrets than competitive baselines.\footnote{This is an extended version of the workshop paper.}
\end{abstract}

\section{Introduction}

Bayesian optimization (BO) is a fundamental and compact framework to study intelligent decision making systems: how to make sequential decisions on data acquisition by reasoning about the Bayesian beliefs on a function, with a goal to optimize this function in a sample-efficient manner. Each decision requires selecting inputs to the function and acquiring their corresponding function evaluations. 

Despite the fact that reasoning with Bayesian beliefs can be hard and costly, BO has found its ways towards real-world deployment because, (1) using a \emph{Gaussian process (GP)} as the Bayesian belief makes inference much cheaper and easier in BO, and (2) global optimization problems on expensive black-box functions need strong sample efficiency and have relatively low requirement on the speed of decision making for data acquisition. BO with GP beliefs have demonstrated numerous successes on such global optimization problems, including hyperparameter tuning in machine learning~\citep{snoek2012practical, wistuba2021few, eggensperger2021hpobench, pineda2021hpob, cowen2022hebo}, action selection in robotics~\citep{driess2017constrained,wang17icra}, search in design spaces of drugs~\citep{pyzer2018bayesian} and chemicals~\citep{shields2021bayesian, griffiths2020constrained}, and even designing aircrafts~\citep{lam2018advances}. However, we can not credit these accomplishments solely to BO, whose efficacy relies heavily on expert knowledge that specifies the initial beliefs on their functions of interest. 

In past literature, researchers have also emphasized the importance of employing a well-specified \emph{GP prior} in BO with both empirical and theoretical evidences~\citep{bogunovic2021misspecified, wang2018regret, kim2019learning}. Using a poorly designed prior, often due to the lack of quantitative understandings of function landscapes, can be detrimental to the performance of BO in applications such as hyperparameter tuning~\citep{schulz2016quantifying}.

How do we obtain well-specified priors for BO? While it is difficult for us to think quantitatively and somehow come up with a reasonable GP out of our mind~\citep{wilson2015human}, we can sort our past experience with related functions and partition the observed data on those functions to employ ``prior learning'' with \textit{i.i.d.} sets of non-\textit{i.i.d.} points~\citep{Baxter1996, minka1997, wang2018regret}. HyperBO~\citep{wang2022pre}, in particular, showed that we can \emph{pre-train} Gaussian process priors on related functions efficiently and effectively for BO applications. 

\cite{wang2022pre}, however, assumed that HyperBO must pre-train on functions with the same domain, i.e. the same BO search space, making it impossible to pre-train on observed data from functions with different domains, let alone generalizing to optimization problems with different search spaces. Essentially, the pre-trained GP prior in HyperBO is restricted to modeling functions that have a fixed input dimension - hence the prior is not operable when we optimize a new function with a different input dimension.

In this work, we aim to pre-train a universal prior that works unanimously for BO with any search spaces. Our new method, HyperBO+, is based on the observations that practitioners commonly use stationary kernels in real-world applications~\citep{balandat2020botorch, Golovin2017, cowen2022hebo} and that the same prior distribution can be shared by some parameters, e.g. length-scales of each input dimension. Even if the function of interest is non-stationary or the output is non-Gaussian, one can employ input or output warping strategies to ensure GPs have the model capacity~\citep{cowen2022hebo}.

In HyperBO+, we partition training data, in the form of past observations, by relevant search spaces and functions in each search space. We then pre-train a hierarchical GP to obtain our universal prior which models functions with different domains. Note that we only need to run the pre-training algorithm once and we can use the same pre-trained hierarchical GP for BO on any functions, as long as their search spaces are relevant to the search spaces in our training data. Theoretically, we show that our pre-trained prior is consistent with the ground-truth distribution of the functions, a fundamental milestone for BO to go beyond the Bayesian cradle. Empirically, we demonstrate the promising results of HyperBO+ to generalize over different search spaces and its competitive performance on classic multi-task problems.

To the best of our knowledge, our work is the first transfer learning approach that is capable of doing optimization on unrestricted search spaces in the literature of BO with GP-based priors. Our contributions include (1) a new two-step pre-training method for hierarchical GPs that can model functions with different domains; (2) HyperBO+, a novel transfer learning BO method that generalizes to universal search spaces; (3) theoretical  and empirical analyses of the asymptotic properties of our pre-training method; and (4) robust results showing the capability of HyperBO+ to generalize to unseen search spaces and its superior BO performance through experiments on synthetic datasets and real-world hyperparameter tuning tasks.

\section{Related work}
Our work is inspired by HyperBO~\citep{wang2022pre}, which showed pre-training helps learning a better GP prior and improves BO performance on different functions with the same domain. Similar ideas~\citep{wistuba2021few,perrone2018scalable, wang2018regret,volpp2020meta} have also been proposed for transfer learning and meta learning in Bayesian optimization but they can only transfer the knowledge among functions with the same domain.  

Recently, \cite{chen2022towards} introduced OptFormer, a transformer based transfer learning method for hyperparameter tuning on universal search spaces, a goal shared by our work. However, due to its core idea of proposing hyperparameters in an end-to-end fashion, OptFormer must be trained on millions of BO trajectories, requires giant transformer models, demands expensive TPU hardware and works only with continuous domains. On the contrary, our model does not have any constraints on how the data is collected, uses simple GP models with a small number of parameters, trains with much shorter time on CPUs and works with both continuous and discrete domains. Due to practical reasons, we do not compare to OptFormer empirically.

Notably, our work focuses on pre-training a universal prior, a critical and specific component of BO. Our pre-training method can be naturally coupled with other important components or heuristics to complete a BO software, e.g. acquisition functions, input and output warping, cross validation etc. Whereas frameworks such as BoTorch~\citep{balandat2020botorch}, Vizier~\citep{Golovin2017} and OptFormer~\citep{chen2022towards} are  hyperparameter tuning software packages. The pre-training method in HyperBO+ can be directly incorporated in BoTorch and Vizier by replacing their default hierarchical GP model.

The theoretical soundness of our approach relies on the quality of the estimated GP parameters during pre-training. The asymptotic behavior for type II maximum likelihood (ML-II) estimators of covariance parameters under a single GP has commonly been studied in two asymptotic frameworks \citep{bevilacqua2019estimation, zhang2005towards}. Under the fixed domain setting, observations are sampled in a bounded set and thus become increasingly dense with more samples. Previous work~\citep{kaufman2013role} shows that only the microergodic parameters can be estimated consistently and provides rigorous proof for isotropic Mat\'ern covariance functions under dimension $\leq 3$. A covariance parameter is microergodic if, when taken two different values, the corresponding Gaussian measures are orthogonal and non-microergodic when the two are equivalent \citep{stein1999interpolation}. 

In the increasing domain setting, observations are collected with a minimal spacing in between data points. Previous results from \cite{mardia1984maximum} and \cite{stein1999interpolation} showed that MLE for covariance parameters are consistent and asymptotically normal in general with mild regularity conditions for increasing domains. 
According to \cite{bachoc2014asymptotic}, the key difference with increasing domain setting is that there is vanishing dependence between observations that are far apart. Thus, larger sample size is more informative of the covariance structure. Our theoretical analysis shows that our problem formulation belongs to the increasing domain setting and previous theoretical guarantees can be used to ensure the performance of our pre-training approach. 

\section{Problem formulation}
\label{sec:problem_formulation}
Our goal is to optimize unseen black-box functions by pre-training on existing data from functions in multiple search spaces, which can have different numbers of dimensions for their respective domains. 

We use the term \textit{super-dataset} to denote all datapoints collected across multiple search spaces, while a \textit{dataset} includes the data from a single search space, and a \textit{sub-dataset} means the collection of datapoints from a single function within a search space. 

More formally, we define a super-dataset as $D =\{ D_i \}_{i=1}^N$. Each dataset $D_i$ consists of observations on a collection of black-box functions $F_i = \{f_{ij}:\mathcal{X}_i\rightarrow\R\}_{j=1}^{M_i}$ where functions in $F_i$ share the same compact search space $\mathcal{X}_i \in \mathbb{R}^{d_i}$. Let $D_i = \{D_{ij}\}_{j=1}^{M_i}$, where each sub-dataset $D_{ij} = \{(x^{ij}_k, y^{ij}_k)\}_{k=1}^{L_{ij}}$. $L_{ij}$ is the number of observations on function $f_{ij}$ perturbed by \textit{i.i.d.} additive Gaussian noise, i.e. $y^{ij}_k \sim \N(f_{ij}(x^{ij}_k), \sigma_{i}^2)$.

For each $i=1,...,N$, we assume all functions in $F_i$ are \textit{i.i.d.} function samples from the same GP: $\mathcal{GP}_i = \mathcal{GP}(\mu_i, k_i)$. 

\begin{figure}
\begin{tikzpicture}
\tikzstyle{main}=[circle, minimum size = 4mm, thick, draw =black!80, node distance = 6mm]
\tikzstyle{para}=[circle, minimum size = 5pt, inner sep=0pt]
\tikzstyle{connect}=[-latex, thick]
\tikzstyle{box}=[rectangle, draw=black!100]
  \node[main, fill = white!100] (theta) [label=below:$\theta_i$] { };
  \node[main] (mu) [right=of theta,label=below:$\mu_i$] {};
  \node[main] (f) [right=of mu,label=below:$f_{ij}$] { };
  \node[main] (k) [below=of mu,label=below:$k_i$] { };
   \node[para, fill = black!100] (alpha) [left=of theta, label=below:$a$] { };
  \node[main, fill = black!10] (y) [right=of f,label=right:$y_{k}^{ij}$] { };
  \node[para, fill = black!100] (x) [above=of y,label=right:$x_{k}^{ij}$] { };
  \node[main, fill = white!100] (sigma) [above=of theta, label=below:$\sigma_i$] { };

  \path (alpha) edge [connect] (theta)
        (alpha) edge [connect] (sigma)
        (sigma) edge [connect] (y)
        (theta) edge [connect] (mu)
        (theta) edge [connect] (k)
		(mu) edge [connect] (f)
		(k) edge [connect] (f)
		(f) edge [connect] (y)
		(x) edge [connect] (y);
  \node[rectangle, inner sep=1mm, fit= (x) (y), label=below right:$L_{ij}$, xshift=-2.4mm, yshift=-6.5mm] {};
  \node[rectangle, inner sep=4.5mm,draw=black!100, fit= (x) (y), xshift=2mm, yshift=-2mm] {};
  \node[rectangle, inner sep=2mm, fit= (x) (y) (f) , label=above left:$M_i$, xshift=6mm, yshift=-3mm] {};
  \node[rectangle, inner sep=6.5mm,draw=black!100, fit= (x) (y) (f) , yshift=-3mm, xshift=2mm] {};
  \node[rectangle, inner sep=0mm, fit= (x) (y) (f) (mu) (k) (sigma) (theta), label=below right:$N$, xshift=7mm, yshift=-5mm] {};
  \node[rectangle, inner sep=8mm,draw=black!100, fit= (x) (y) (f) (mu) (k) (sigma) (theta), yshift=-2mm, xshift=2mm] {};
\end{tikzpicture}
\centering
\caption{Graphical model for a hierarchical GP.}
\label{fig:gpgraph1}
\vspace{-10pt}
\end{figure}
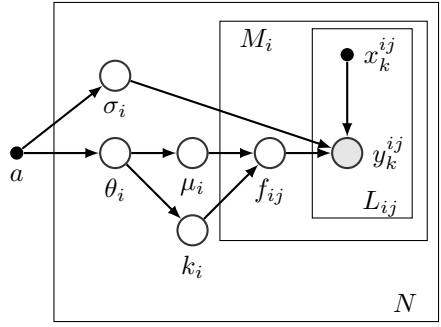

For each function set $F_i$ and its corresponding $\mathcal{GP}_i$ with mean function $\mu_i: \mathcal{X}_i \rightarrow \mathbb{R} $ and kernel $k_i: \mathcal{X}_i \times \mathcal{X}_i \rightarrow \mathbb{R}$, we denote the parameters of $\mu_i, k_i$ by $\theta_i$. 

Our major assumption is that the GP parameters and noise standard deviation $\{(\theta_i, \sigma_i)\}_{i=1}^{N}$ are \textit{i.i.d.} samples from a distribution, as in $(\theta_i, \sigma_i) \sim p((\theta, \sigma); a)$, which is parameterized by $a$. The distribution $p((\theta, \sigma); a)$ is a \textit{universal prior} for all search spaces. Fig.~\ref{fig:gpgraph1} illustrates the graphical model of our hierarchical GP.

For simplicity, we use a length-scale based kernel, e.g. Mat\'ern kernel, and assume that each dimension of the GP parameters and the noise variance are all independent. We further assume that all length-scale parameters share the same prior distribution. Thus, we can estimate the distribution $p(\cdot; a)$ using the super-dataset and this learned universal prior can be used to guide BO on new black-box functions, either from seen search spaces or from unseen search spaces.

\paragraph{Evaluation metrics.}
    For an unseen blackbox ``testing function'' $f$, we run BO on the function for $T$ steps and accumulate the observations $\{x_t, y_t\}$. We evaluate the optimization performance of a given method on an individual testing function using the \textit{normalized simple regret} 
    \begin{align} 
    r_T = \frac{\mathrm{max}_{x \in \mathcal{X}} f(x) - \mathrm{max}_{t \in [T]} y_t}{\mathrm{max}_{x \in \mathcal{X}} f(x) - \mathrm{min}_{x \in \mathcal{X}} f(x)}.
    \end{align}
    We use the \textit{mean of normalized simple regret} on all testing functions as the overall evaluation metric.\\
    Additionally, we also report the \textit{negative log-likelihood (NLL)} of the compared models (either a hierarchical GP or a vanilla GP) on the training and testing datasets. For a vanilla GP parameterized by $(\theta, \sigma)$, the NLL on a dataset $D$ is defined as $\mathrm{NLL}(\theta, \sigma) = -\log p(D | \theta, \sigma)$. For a hierarchical GP parameterized by $a$, the NLL is defined as follows and is computed via sampling in practice. 
    \begin{equation}
    \begin{split}
    \label{equ:estimate-nll}
         &\mathrm{NLL}(a) = -\log p(D \mid a)\\ &= -\log \left( \int_{(\theta, \sigma)} p(D \mid \theta, \sigma) p((\theta, \sigma); a)
    \mathrm{d}(\theta, \sigma)\right) \\
    &\approx -\log \left( \frac{1}{Q} \sum_{q=1}^{Q} p(D \mid \theta_q, \sigma_q) p((\theta_q , \sigma_q); a)\right)
    \end{split}
    \end{equation}
    where $\{(\theta_q, \sigma_q)\}_{q=1}^{Q}$ are \textit{i.i.d.} samples from $p((\theta, \sigma); a)$.

\section{Our method}
The HyperBO+ framework we propose consists of mainly two phases: (1) Training: estimate the universal prior $a$ from the super-dataset $D = \{D_i\}_{i=1}^N$ with a two-step approach. (2) Optimization: running BO with the hierarchical GP parameterized by the learned $a$ on testing functions. 

\subsection{Two-step pre-training}
\label{ssec:twostep}
\paragraph{Estimating GP parameters of each search space.} For each function collection $F_i$ with domain $\mathcal{X}_i$, we can infer its GP parameters $\theta_i$ and noise standard deviation $\sigma_i$ by minimizing the negative log-likelihood of the dataset as in the original HyperBO~\citep{wang2022pre}:
\begin{align}
\label{eq:nll}
\begin{split}
   & L(\theta,\sigma \mid D_i) = -\log p (D_i \mid \theta, \sigma) \\
   & = -\sum_{j = 1}^{M_i} \log p(D_{ij} \mid \theta, \sigma) \\
   & = \sum_{j=1}^{M_i} \left( \frac{1}{2} Y_{ij}^{\intercal} K^{-1} Y_{ij} + \frac{1}{2}\log |K| + \frac{L_{ij}}{2} \log 2\pi \right) 
\end{split}
\end{align} 
where $\bm{x}^{ij} = [x_k^{ij}]_{k=1}^{L_{ij}}, \bm{y}^{ij} = [y_k^{ij}]_{k=1}^{L_{ij}}$, $Y_{ij}=\left(\bm{y}^{ij} - \mu(\bm{x}^{ij})\right)$, $K = k(\bm{x}^{ij}) + \sigma^2 \bm{I}$, and $\mu$ and $k$ are the mean and covariance functions parameterized by $\theta$.
Specifically, we minimize the negative log-likelihood to get the maximum likelihood estimator (MLE) of $\theta_i, \sigma_i$ as
\begin{align}
(\hat{\theta}_i, \hat{\sigma}_i)_{ML} = \argmin_{\theta, \sigma} L(\theta, \sigma, D_i).
\end{align} 
\paragraph{Estimate the universal prior.} Using the estimated $\{(\hat{\theta}_i, \hat{\sigma}_i)\}_{i=1}^{N}$ from all datasets, we can use the the maximum likelihood estimator for the universal prior parameter $a$ as $\hat{a} = \argmax_{a} p(\{(\hat{\theta}_i, \hat{\sigma}_i)\}_{i=1}^{N}; a)$. For instance, we can choose the universal prior to be a normal distribution, i.e. 
\[
p(\{(\hat{\theta}_i, \hat{\sigma}_i)\}_{i=1}^{N}; a) = \N(a_0, \textbf{I} a_1)
\]
where $a = [a_0, a_1] \in \R^2$, and directly estimate the mean and variance as the parameter $a$. For length-scales, we treat all estimated length-scales as \textit{i.i.d.} samples from the same distribution. 
 \begin{algorithm}[H]
       \caption{HyperBO+ pre-training and Bayesian optimization with acquisition function~$ac(\cdot)$.}\label{alg:hyperbo}
  \begin{algorithmic}[1]
    \Function{HyperBO+\,}{$f, D$}
    \For{$D_i \in D $} \label{alg:pre-trainstart}
        \State $\hat{\theta}_i, \hat{\sigma}_i \gets \textsc{Pre-Train}(D_i)$\label{alg:train}
    \EndFor \label{alg:pre-trainend}
    \State $\hat a \gets \textsc{MLE}(\{\hat{\theta}_i, \hat{\sigma}_i\}_{i=1}^{N})$ \label{alg:mle}
    \State $D_f \gets \emptyset$
    \For{$t = 1,\cdots, T $} \label{alg:bostart}
        \State $x_t\gets \underset{x\in\mathfrak X}{\argmax}{\,ac_t\left(x; \hat{a}) \right)}$ (Eq.~\ref{equ:ac_fun})\label{alg:strategy}
        \State $y_t\gets$ \textsc{Observe}$\left(f(x_t)\right)$
        \State $ D_f \gets D_{f}\cup \{(x_t,y_t)\}$
      \EndFor \label{alg:boend}
     \State \Return $D_f$
    \EndFunction
  \end{algorithmic}
\end{algorithm}
\subsection{Bayesian optimization}

HyperBO+ models functions via a hierarchical GP with pre-trained universal prior parameter $\hat{a}$. At step $t$ of Bayesian optimization when optimizing a testing function $f$, we compute the acquisition function for HyperBO+ by first computing the posterior distribution of GP parameters and noise stand deviation $(\theta, \sigma)$ themselves, then taking the average of acquisition function values according to $(\theta, \sigma)$-s sampled from this posterior distribution:
\begin{align}
\label{equ:ac_fun} \hspace{-0.2cm}
ac_t(x; \hat{a}) = \sum_{r=1}^{R} \left[ac_t(x; \theta_r, \sigma_r) p((x_k, y_k)_{k=1}^{t} \vert \theta_r, \sigma_r)\right]
\end{align}
where $(\theta_1, \sigma_1), \dots, (\theta_R, \sigma_R)$ are \textit{i.i.d.} samples from the prior distribution $p((\theta, \sigma); \hat{a})$. By injecting the term $p((x_k, y_k)_{k=1}^{t} \vert \theta_r, \sigma_r)$ to the acquisition function value of each sample, we convert sampling from the prior distribution to sampling from the posterior distribution $p((\theta, \sigma)  (x_k, y_k)_{k=1}^{t}; \hat{a})$ according to Bayes' Rule $p((\theta, \sigma) \vert (x_k, y_k)_{k=1}^{t}; \hat{a}) = p((x_k, y_k)_{k=1}^{t} \vert \theta_r, \sigma_r) p((\theta_r, \sigma_r); \hat{a})/ p((x_k, y_k)_{k=1}^{t})$. Here we can ignore the denominator $p((x_k, y_k)_{k=1}^{t})$ since the BO selection is invariant under any positive affine transformation of the acquisition function. Note that posterior sampling could be a preferred option than re-weighting with likelihood. We find our re-weighting strategy to be sufficient since the samples from the pre-trained prior are often representative of the samples from the posterior.

\section{Theoretical analysis}
\label{sec:theory}
\begin{figure*}
\centering
    \includegraphics[width= 0.7\linewidth]{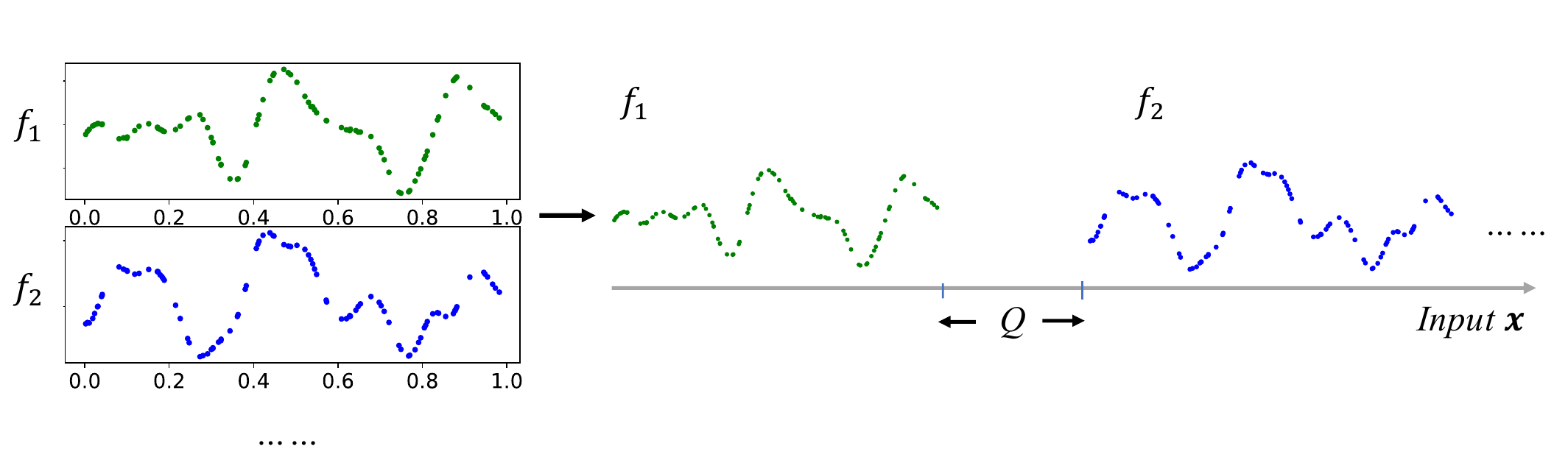}
    \caption{Converting a series of sub-datasets equivalently to samples from a single function with increasing observation locations.}
    \label{fig:gp-proof}
\end{figure*}

Showing the asymptotic behaviors of the estimated GP parameter $\hat\theta_i$ as the number of datasets and sub-datasets approaches infinity is non-trivial. 
These asymptotics differs from standard analyses of estimators where samples are drawn \textit{i.i.d.} from the same distribution. Oftentimes, the determinant $|K|$ and gram matrix inverse $K^{-1}$ in the likelihood function has no explicit expressions \citep{bachoc2021asymptotic} thus making it even more challenging to analyze the properties of the estimated GP parameter based on type II maximum likelihood.

We recall that in the increasing domain setting, there is a minimal spacing between any given pairs of observations and thus make the sampling domain unbounded as the number of observations increases. That is, there exists a fixed $\Delta > 0$ such that 
\begin{align}
    \text{inf}_{i,j \in \mathbb{N}, i\neq j} \| x_i - x_j\| \geq \Delta. \label{eq:increasing}
\end{align}

In previous work showing the consistency of ML estimators for GP parameters, they consider a a single function sampled from a zero-mean GP with $n$ observations to minimize the NLL
\begin{align*}
    L_n(\theta) = \frac{1}{n} \log (|K_\theta|) + \frac{1}{n} y^T K_{\theta}^{-1} y
\end{align*} for $\hat{\theta}_{ML} \in \argmin_{\theta \in \Theta} L_n(\theta)$, where $K_{\theta}$ is the covariance matrix parameterized by $\theta$. 
The authors also showed that though the MLE given finite observations may not be unique, it converges to the unique global minimizer with probability converging to one as the number of observations goes to infinity. Extensions of such results on increasing domain for GP with non-zero mean function were discussed in \cite{bachoc2020asymptotic}.

We now analyze the asymptotic behavior of the first step in \S\ref{ssec:twostep} - GP parameter estimation for each search space. 
We make the following \textbf{assumptions}:
\setenumerate[1]{label=(\arabic*)}
\begin{enumerate}[leftmargin=0.5 cm]
    \item Each GP is zero-mean with known noise variance $\sigma^2$. Its ground-truth kernel belongs to the Mat\'ern covariance family with known smoothness term $\nu$ (e.g. $\nu = 3/2$) and the parameter $\theta$ consists of the length-scale only.
    
    \item The kernel for each GP is fixed and can be parameterized by a unique $\theta_0 \in \Theta$ for some compact $\Theta$. Moreover, the setup is \textit{well-specified}, where the true covariance parameters $\theta_0$ do belong to the search space $\Theta$ used for estimation.
\end{enumerate}

In our problem setup, a dataset $D_i = \{D_{ij}\}_{j=1}^{M_i}$ ($i = 1, \dots, N$) contains a finite number of observations in each of its sub-datasets. For the following analysis we assume that there exists a fixed $\delta > 0$ such that $\|x^{ij}_k - x^{ij}_{k'}\| \ge \delta, (k \ne k')$ for all sub-datasets $D_{ij}$. As the number of sub-datasets $M_i \rightarrow \infty$, we can show that it is equivalent to the increasing domain setting within a single function from GP and prior work guarantee good asymptotic properties of the ML estimation for covariance parameters \citep{mardia1984maximum, stein1999interpolation}.

\begin{lem}
\label{prop:increasing} 

Given assumption (1), for any $i\in \{1,\cdots, N\}$, $\GP_i$ and its corresponding \emph{dataset} $D_i = \{D_{i1}, \cdots, D_{iM_i}\}$, there exists a sub-dataset $\bar D$ observed on a function $f'\sim \GP_i$, such that $\bar D$ satisfies the increasing domain characterization in Eq.~\ref{eq:increasing}, and
\begin{align} \label{eq:prop}
    L(\theta \mid D_i) \equiv L(\theta \mid \{\bar D\}),
\end{align}
where $L(\theta \mid \mathcal D) = - \sum_{ \tilde D \in \mathcal D}\log p(\tilde D \mid \theta).$
\end{lem}

Intuitively, this lemma shows that observations from multiple independently generated functions from a fixed GP can be viewed as being sampled from one function, with infinitely large interval between sub-datasets' observations. We illustrate this process with two sub-datasets on a 1-dimensional domain in Fig.~\ref{fig:gp-proof}. A detailed proof is presented in \S\ref{app:theory}.

We introduce our \textbf{assumption (3)} that there is sufficient information in sampling locations $x_1, x_2, \dots, x_n$ of the augmented sub-dataset $\bar D$ to distinguish covariance functions $k_\theta$ with $k_{\theta_0}$~\citep{bachoc2014asymptotic}. That is, for $\theta \in \Theta$
    \begin{align}
         \resizebox{.93\hsize}{!}{$\liminf_{n \rightarrow \infty} \inf_{\|\theta - \theta_0\| \geq \epsilon} \frac{1}{n} \sum_{i \neq j} (k_\theta(x_i - x_j)) - k_{\theta_0}(x_i - x_j))^2 > 0$}. \tag{Asymptotic Identifiability}
    \end{align} 

\begin{thm}
\label{theorem3}
Given assumptions (1)-(3),
for dataset $D_i$ with $M_i$ sub-datasets generated from the same mean and covariance function, as $M_i \rightarrow \infty$, we have
$    \hat{\theta}_{ML} \rightarrow^p \theta_0.$
\end{thm}

\begin{thm}
\label{theorem4}
Given assumptions (1)-(3), as the number of datasets, and sub-datasets $N, M_i \rightarrow \infty, \forall i$, the maximum likelihood estimator for the prior distribution $a$ is consistent.
\end{thm}
Theorem~\ref{theorem3} and~\ref{theorem4} are easy to show and the proofs can be found in \S\ref{app:theory}.

\begin{remark} (\cite{ye2017closed} and \cite{rice2006mathematical})\\
(1). When the prior is assumed to be from a Gamma distribution $\Gamma(\alpha, \beta)$, the MLE $\hat{\alpha}, \hat{\beta}$ is consistent and asymptotically normal as the number of samples $n \rightarrow \infty$ with distribution $\mathcal{N}((\alpha, \beta), \frac{1}{n} \mathcal{I}(\alpha, \beta)^{-1})$, where $\mathcal{I}$ is the Fisher information matrix. \\
(2). Similar results hold when the prior is a normal distribution parametrized by $\mu, \sigma$: as $n \rightarrow \infty$, we have $\hat{\mu}_{ML}, \hat{\sigma}_{ML} \rightarrow \mu, \sigma$.

\end{remark}

\begin{figure}[t]
    \centering
    \begin{subfigure}{0.55\textwidth}
        \includegraphics[width=\linewidth]{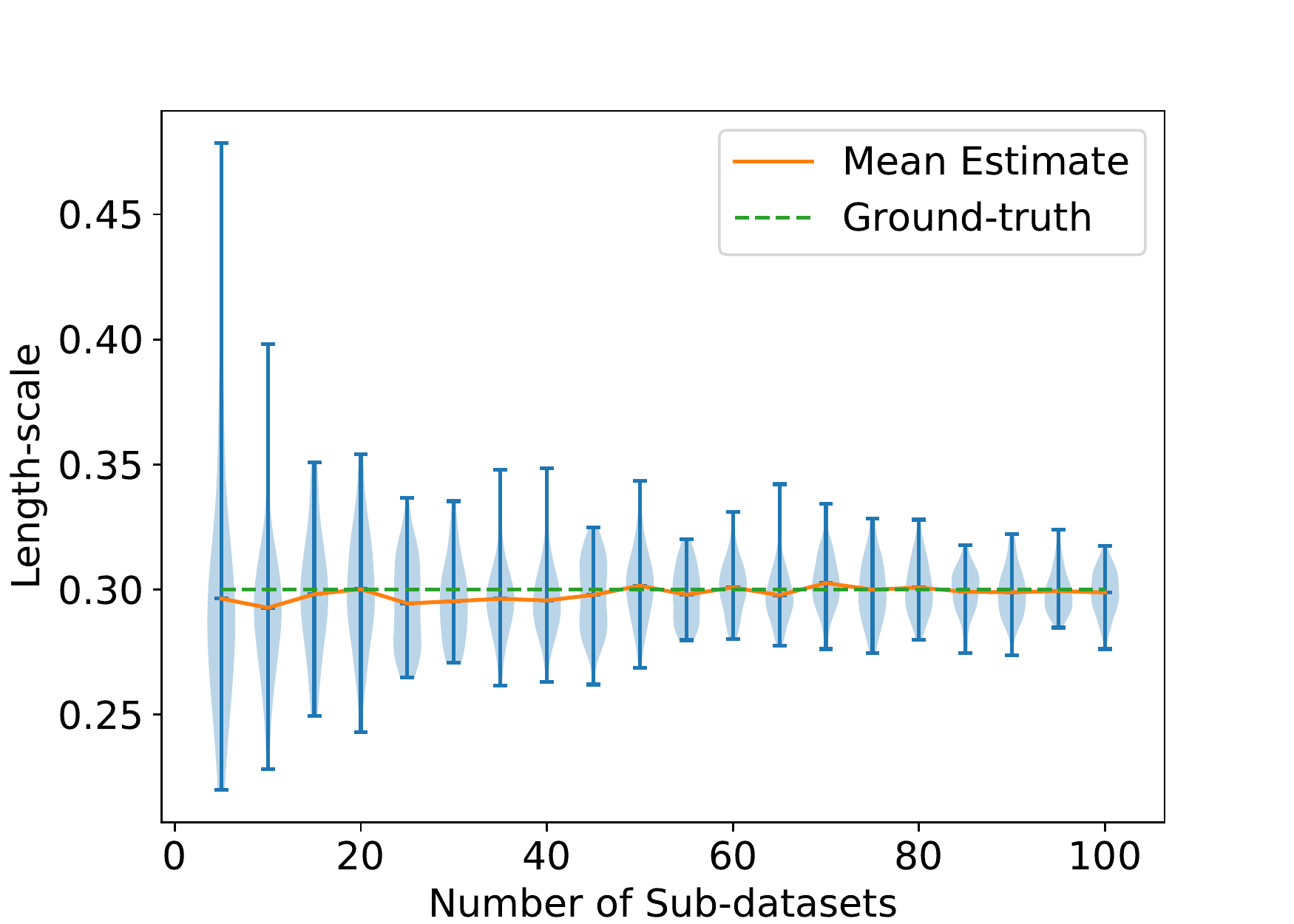}
    \end{subfigure}
    \caption{Estimated length-scale parameter (distribution plotted over 50 random runs) of a 1-dimensional GP as the number of sub-datasets increases. Each sub-dataset has 25 observations. 
    The variance of estimated length-scale parameter decreases as the number of sub-datasets increases.
    }
    \label{fig:one-gp-asymptotic-lengthscale}
\end{figure}

A key observation from our theoretical analyses is that with sufficient observations in each dataset $D_i$, increasing the number of sub-datasets can effectively improve the estimation of the covariance parameters and thus the prior. Fig.~\ref{fig:one-gp-asymptotic-lengthscale} demonstrates the asymptotic behavior empirically for fitting the length-scale parameter of a single GP using an increasing number of sub-datasets. Note that while our analysis considers only the length-scale estimation, empirical results show strong evidence for the consistency of function mean and signal/noise variance as well, which is illustrated in Fig.~\ref{fig:one-gp-asymptotic-constant-sigvar-noisevar} in the supplement. Empirical asymptotic behavior of the two-step pre-training method is demonstrated in \S\ref{subsec:empirical-asymptotic-behaviors}. 

\section{Experiments}
\label{sec:experiments}


We demonstrate the performance of HyperBO+ on two datasets: a synthetic Super-dataset and HPO-B~\citep{pineda2021hpob}, a collection of real-world hyperparameter tuning tasks that involves multiple search spaces. We generate the synthetic super-dataset with a known ground-truth prior distribution over GP parameters, which is helpful for comparing with the ground-truth and empirically demonstrating asymptotic behaviors of the two-step pre-training method. 
The same cannot be done for real-world datasets since there is not a known ground-truth prior.

\textbf{Synthetic Super-dataset} is generated over multiple search spaces following the generative process illustrated in Fig.~\ref{fig:gpgraph1}. The super-dataset includes 20 datasets (each with its own search space) with 10 sub-datasets in each dataset. Each sub-dataset includes noisy observations at 300 random input locations in its respective search space. The dimensions of search spaces are between 2 and 5.

\textbf{HPO-B Super-dataset}~\citep{pineda2021hpob} is a large-scale benchmark for hyperparameter optimization. As a super-dataset, it consists of 16 different search spaces and more than 6 million evaluations in total. The dimensions of these search spaces vary from 2 to 18.

\subsection{Experiment setups}

To test the capability of HyperBO+ to generalize to new tasks in both seen and unseen search spaces, we design experiments for the two super-datasets and compare HyperBO+ (and its variants) with 5 competitive baselines including HyperBO~\citep{wang2022pre}, which is known to be the state-of-the-art prior pre-training method for BO.

For the Synthetic Super-dataset, we design the following two setups as two different ways of splitting training and testing data.

\textbf{Setup A} \; is designed to demonstrate the ability of HyperBO+ to generalize its learned prior to unseen search spaces, potentially with different input dimensions. We split the Synthetic Super-dataset into training datasets and testing datasets. In particular, we use 80\% of the super-dataset for training and the remaining 20\% for testing. To test BO performances of different methods, we report their average normalized simple regrets on sub-datasets in the testing datasets. We also report the NLLs of different GP-based methods on training datasets and testing datasets.

\textbf{Setup B} \; aims to to test the ability of HyperBO+ to generalize to new functions in seen search spaces so that we can compare its performance with HyperBO. We split each dataset in the Synthetic Super-dataset into training sub-datasets and testing sub-datasets. We use 80\% of each dataset as the training set of sub-datasets and the remaining 20\% for testing. We report the average normalized simple regrets on all testing sub-datasets and the NLLs of different GP-based methods on training and testing sub-datasets. HyperBO individually pre-trains one GP per search space, while HyperBO+ learns a universal hierarchical GP prior by pre-training on the collection of training sub-datasets of all search spaces and uses the same pre-trained prior to run BO on testing sub-datasets for all search spaces.

\textbf{HPO-B Setup}\; HPO-B Super-dataset comes with a pre-specified train/test split: each dataset (search space) contains multiple training sub-datasets and multiple testing sub-datasets. For BO performance, we report the average normalized simple regrets on testing sub-datasets in all search spaces. We also report the NLLs of different GP-based methods on training sub-datasets and testing sub-datasets. HyperBO works by individually pre-training a single GP per search space. On the other hand, HyperBO+ learns a universal hierarchical GP prior by pre-training on the collection of all training sub-datasets.

\textbf{Baselines:} \; (1) Random sampling for optimization. (2) A hand-specified (and potentially misspecified) hierarchical GP prior, fixed over all search spaces. (3) A non-informative hierarchical GP prior, fixed over all search spaces. (4) HyperBO~\citep{wang2022pre}. Note that HyperBO is only applicable to Setup B. (5) The ground-truth hierarchical GP prior, which is available to Synthetic Super-dataset. Please see \S\ref{app:exp_setups} for the exact configurations, including GP prior parameters, of all baselines.

\textbf{Variants of HyperBO+:} \; In addition to the standard HyperBO+, we also test the performance of the following two variants. (1) Instead of fitting a parameterized universal prior with MLE, $^{\mathsf{x}}$HyperBO+ uses the uniform distribution over the finite set of GP parameter values $\{(\hat{\theta}_i), \hat{\sigma_i}\}_{i=1}^N$ as an explicit prior distribution (i.e. crossing out the MLE step at Line~\ref{alg:mle} of Alg.~\ref{alg:hyperbo}).
(2) For the HPO-B Super-dataset, instead of pre-training HyperBO+ on training sub-datasets of all search spaces, $^{\mathsf{z}}$HyperBO+ is pre-trained with the training sub-datasets that do not belong to the search space of the testing sub-dataset (i.e. zero knowledge of its own search space). $^{\mathsf{z}}$HyperBO+ is the same as the standard HyperBO+ in Setup A of the Synthetic Super-dataset. 

We fit the GP parameters in each search space by minimizing Eq.~\ref{eq:nll} using L-BFGS~\citep{liu1989lbfgs} for the Synthetic Super-dataset, and using the Adam optimizer~\citep{kingma2015adam, wistuba2021few} for HPO-B. All the GP-based methods use a constant mean function and a Mat\'ern kernel with smoothness parameter $\nu=3/2$. More experiment details and analyses can be found in \S\ref{app:exp_setups} and \S\ref{app:more_exp}.

\subsection{Empirical asymptotic behaviors}
\label{subsec:empirical-asymptotic-behaviors}

\begin{figure}[t]
    \centering
    \begin{subfigure}{0.55\textwidth}
        \includegraphics[width=\linewidth]{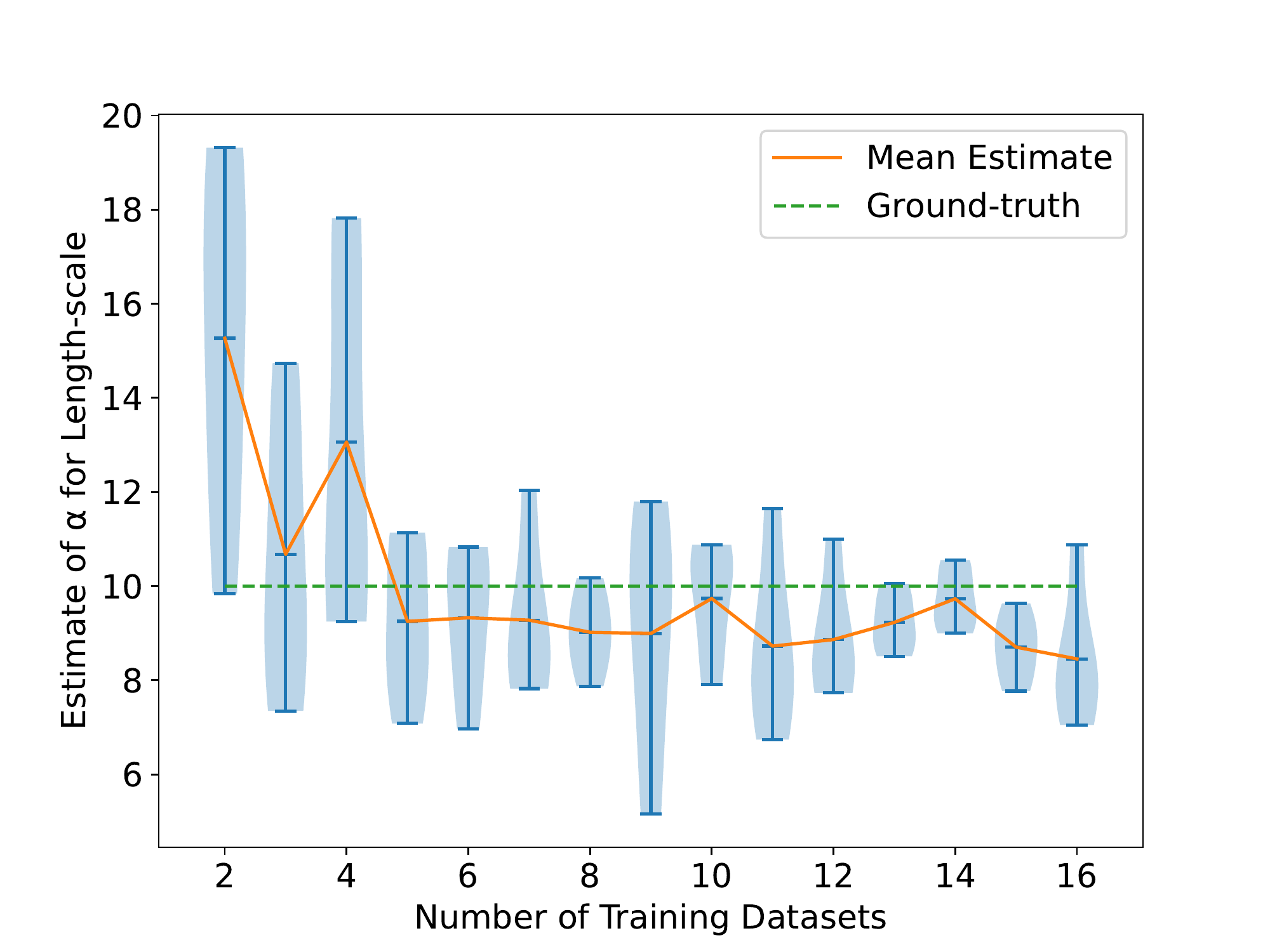}
    \end{subfigure}
    \caption{Estimated shape parameter $\hat{\alpha}$ (y-axis) of Gamma distribution prior for the length-scale GP parameter w.r.t. the number of training datasets (x-axis). We show the mean and violin plots for $\hat{\alpha}$ over 5 random seeds.}
    \label{fig:two-step-asymptotic-lengthscale-synthetic}

\end{figure}

\begin{figure}[t]
\vspace{-3pt}
    \centering
        \includegraphics[width=0.55\textwidth]{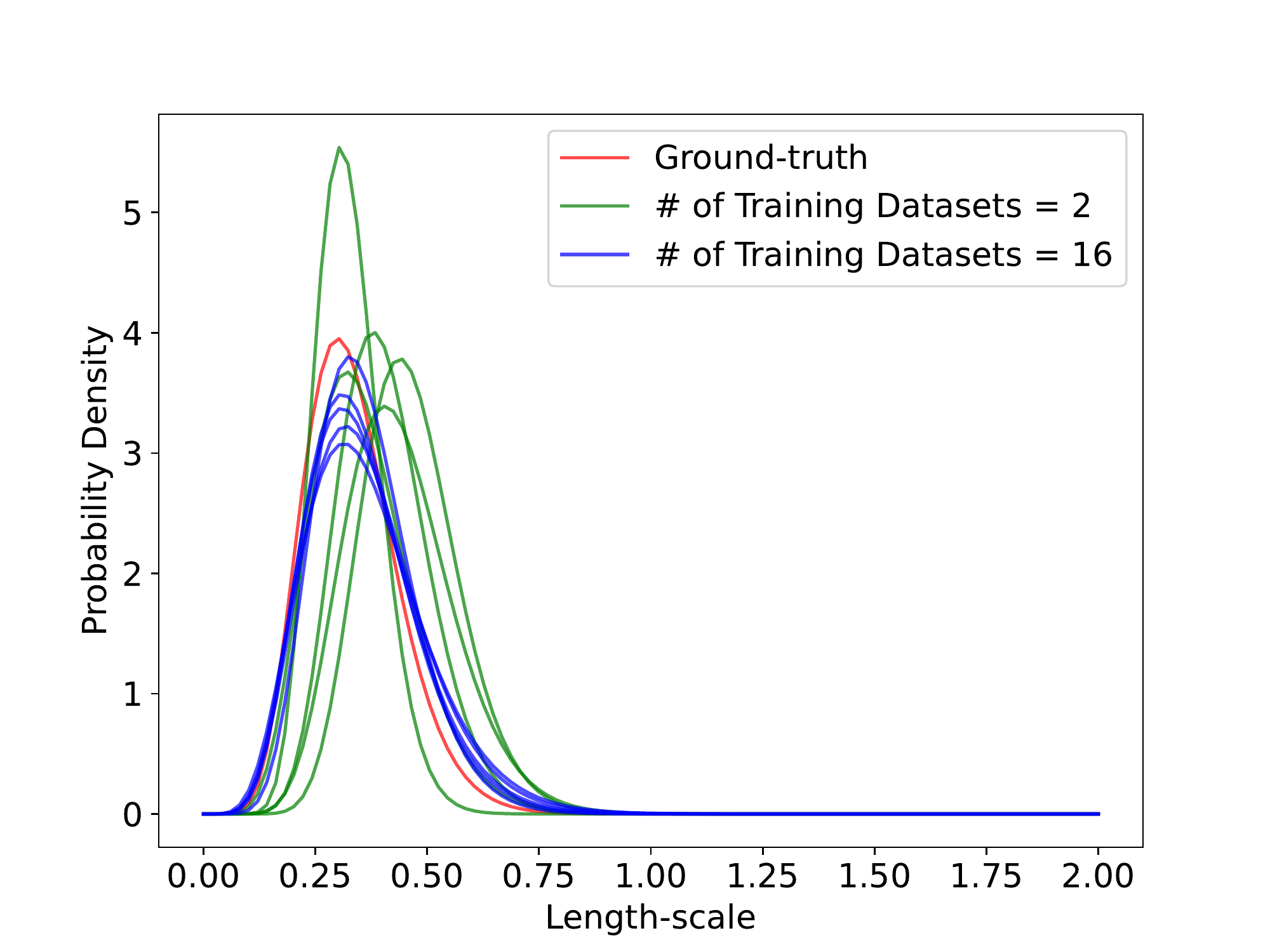}
        \caption{We plot the probability density functions of the Gamma distributions that model length-scales. The pre-trained Gamma distributions with 16 training datasets are more stable than those with 2 training datasets and match well with the ground-truth.}
        \label{fig:two-step-asymptotic-lengthscale-synthetic-distribution-main}
\end{figure}

\begin{figure}[t]
    \centering
    \begin{subfigure}{0.55\textwidth}
        \includegraphics[width=\linewidth]{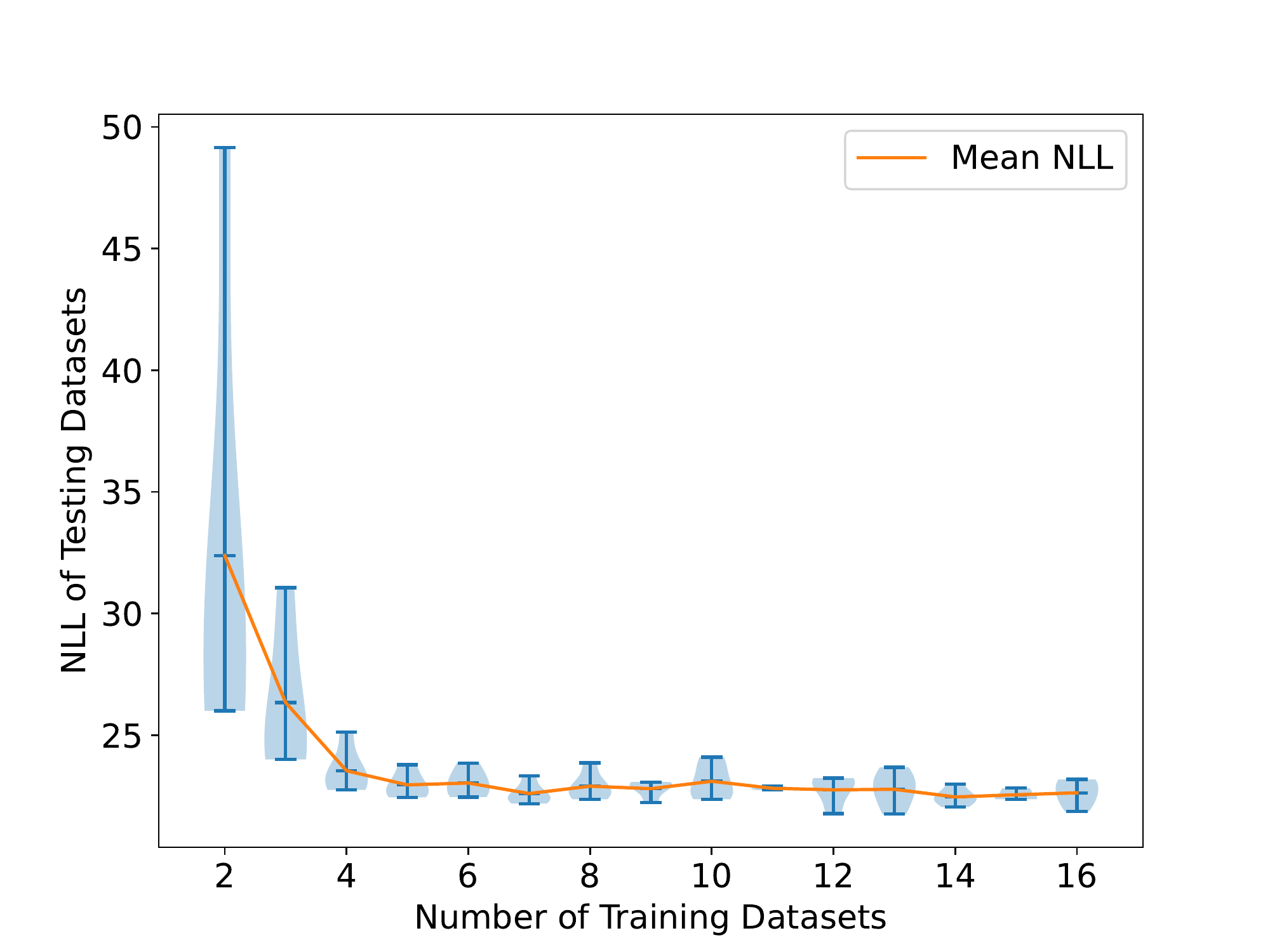}
    \end{subfigure}
    \caption{NLLs of the testing datasets on pre-trained priors as the number of training datasets increases. We show the mean and violin plots for the NLLs over 5 random seeds.}
    \label{fig:two-step-asymptotic-test-nll-synthetic}

\end{figure}

To empirically demonstrate asymptotic properties of our two-step pre-training method analyzed in \S\ref{sec:theory}, we present results with increasing number of training datasets used in Setup A on Synthetic Super-dataset. The testing super-dataset is fixed, while the training datasets are uniformly randomly sampled from all available training datasets. We repeat the experiments 5 times with different random seeds.  

Fig.~\ref{fig:two-step-asymptotic-lengthscale-synthetic} shows the estimated shape parameter $\alpha$ of the Gamma distribution prior $\Gamma(\alpha, \beta)$ which models the length-scale parameter of the GP for each training dataset. 
We illustrate how well the pre-trained Gamma prior matches the ground-truth by plotting the probability density functions in Fig.~\ref{fig:two-step-asymptotic-lengthscale-synthetic-distribution-main}. 

We observe that as the number of training datasets increases, the variance of the estimated $\hat{\alpha}$ gradually decreases and the mean becomes closer to the ground-truth in Fig.~\ref{fig:two-step-asymptotic-lengthscale-synthetic}. Similarly, Fig.~\ref{fig:two-step-asymptotic-lengthscale-synthetic-distribution-main} also shows that more training datasets help the stability of pre-training. This is consistent with our theoretical analysis of asymptotic properties of the two-step pre-training method. The mean value of $\alpha$ is still a bit off from the ground-truth, which is expected given that there are only 16 samples for the second step of pre-training. However, this estimation quality is good enough for HyperBO+ to outperform baselines on this Synthetic Super-dataset as shown below in \S\ref{subsec:bo-results}. 

In Fig.~\ref{fig:two-step-asymptotic-test-nll-synthetic}, we show the NLL of the testing datasets on the estimated hierarchical GP prior  as a function of the number of training datasets. Both the mean and variance of the NLL on testing dataset drop as the number of training datasets increases, which is consistent with our asymptotic analysis. 

\subsection{Bayesian optimization results}

\begin{figure}[h]
    \centering
    \includegraphics[width= \linewidth]{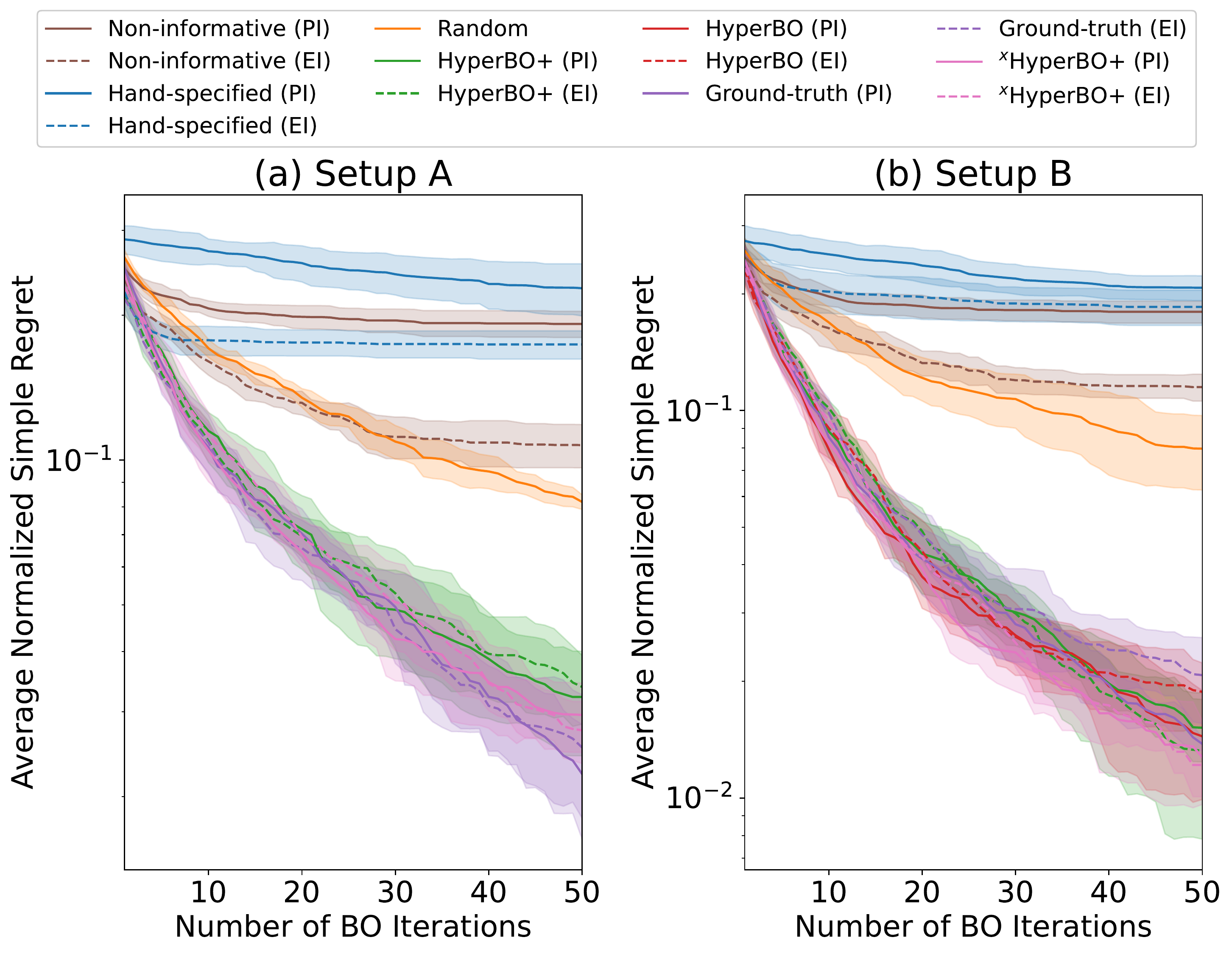}
    \caption{Average normalized simple regret of each method during BO in both Setup A and Setup B for the Synthetic Super-dataset. The averages are taken over 5 random seeds, and the highlighted areas show $\mathrm{mean} \pm \mathrm{std}$ for each method. Results for acquisition functions PI and EI are shown here.}
    \label{fig:bo-regrets-eipi-synthetic}
\end{figure}

\begin{figure}[h]
    \centering
    \includegraphics[width= 0.9\linewidth]{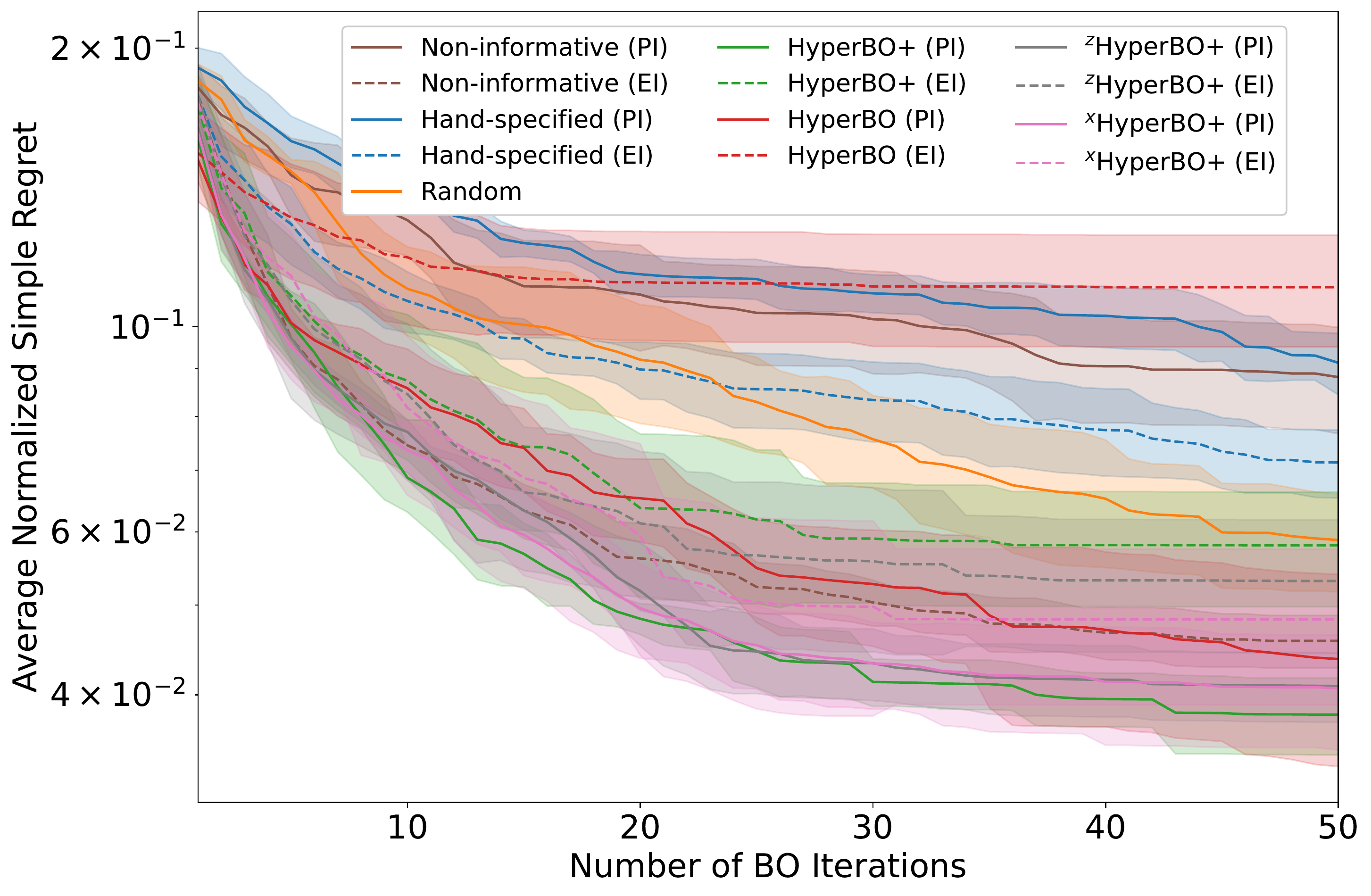}
    \caption{Average normalized simple regret of each method during BO for the HPO-B Super-dataset. The averages are taken over 5 random seeds, and the highlighted areas show $\mathrm{mean} \pm \mathrm{std}$ for each method. Results for acquisition functions PI and EI are shown here.}
    \label{fig:bo-regrets-eipi-hpob}
\end{figure}

\begin{figure*}[!htbp]
    \centering
    \includegraphics[width=\linewidth]{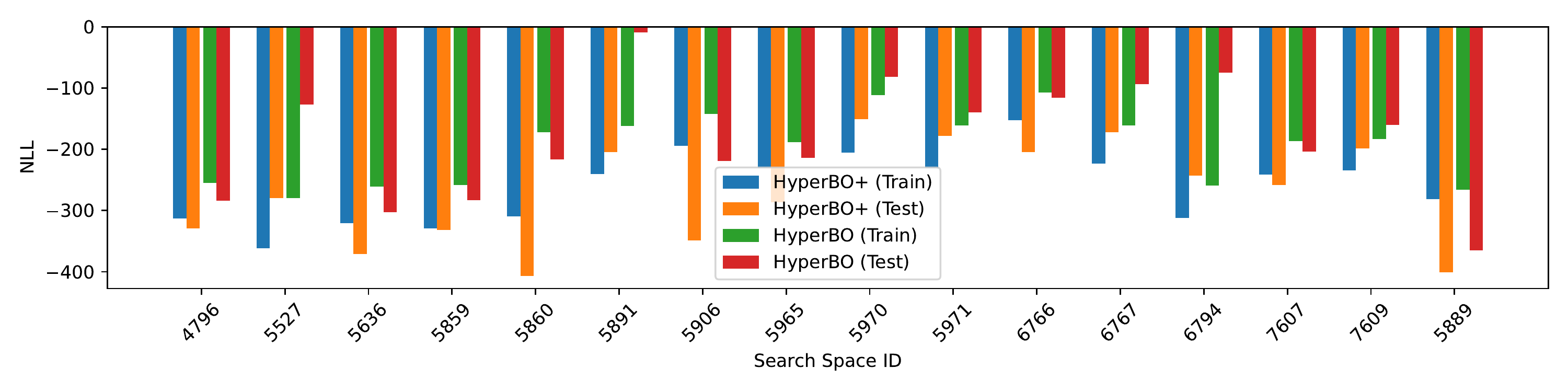}
    \caption{NLL evaluations of training/testing data on HyperBO+ and HyperBO for the HPO-B Super-dataset.}
    \label{fig:nll-per-search-space-hpob}
\end{figure*}

\begin{figure}[h]
    \centering
    \includegraphics[width=0.6\linewidth]{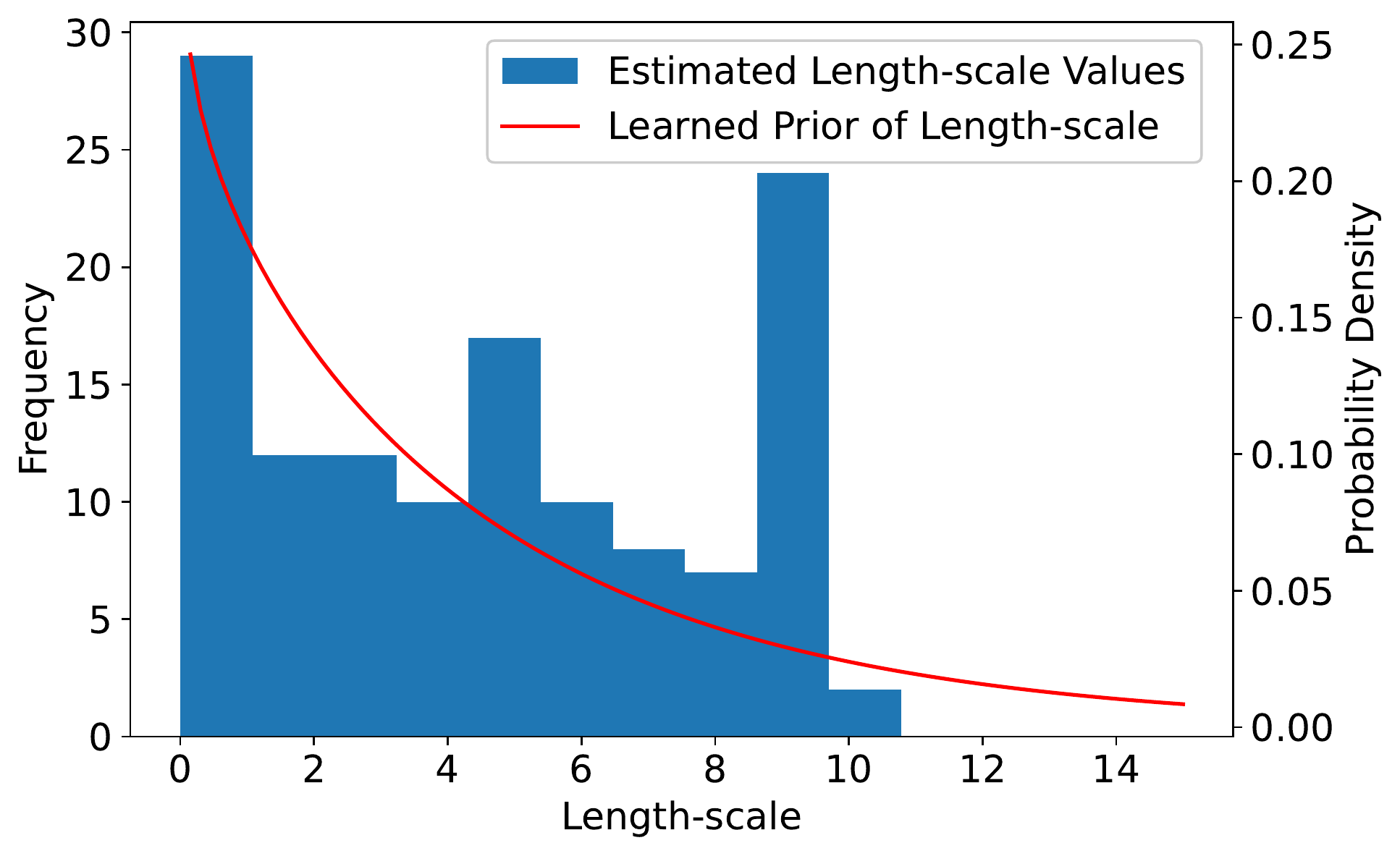}
    \caption{The estimated length-scale values for each dimension of every search space and the learned Gamma prior distribution of length-scale for HPO-B Super-dataset.}
    \label{fig:lengthscale-dist-hpob}
\end{figure}


\label{subsec:bo-results}

\begin{table*}[!htpb]
\caption{NLL evaluations of the training and testing data on different models and normalized simple regret of running BO with each method (PI acquisition function) in both Setup A and Setup B for the Synthetic Super-dataset. ($\mathrm{mean} \pm \mathrm{std}$)}
\label{tab:nll-regret-synthetic}
\begin{center}
\begin{adjustbox}{width=\textwidth}
\begin{tabular}{c|lll|lll}
\hline
\multirow{2}{*}{Methods} & \multicolumn{3}{c|}{Setup A}                                                                          & \multicolumn{3}{c}{Setup B}                                                                          \\ \cline{2-7} 
                         & \multicolumn{1}{c|}{Train NLL} & \multicolumn{1}{c|}{Test NLL} & \multicolumn{1}{c|}{BO Regret} & \multicolumn{1}{c|}{Train NLL} & \multicolumn{1}{c|}{Test NLL} & \multicolumn{1}{c}{BO Regret} \\ \hline
Non-informative          & \multicolumn{1}{l|}{$122.2 \pm 0.4$}             & \multicolumn{1}{l|}{$106.7 \pm 0.2$}            &  \multicolumn{1}{l|}{$0.192 \pm 0.01$}                               & \multicolumn{1}{l|}{$128.8 \pm 0.3$}             & \multicolumn{1}{l|}{$127.0 \pm 0.7$}            &   \multicolumn{1}{l}{$0.180 \pm 0.01$}                             \\ \hline
Hand-specified           & \multicolumn{1}{l|}{$99.9 \pm 0.4$}             & \multicolumn{1}{l|}{$52.2 \pm 0.6$}            &    \multicolumn{1}{l|}{$0.228 \pm 0.03$}                             & \multicolumn{1}{l|}{$90.5 \pm 0.6$}             & \multicolumn{1}{l|}{$91.5 \pm 1.0$}            &     \multicolumn{1}{l}{$0.207 \pm 0.02$}                           \\ \hline
HyperBO                  & \multicolumn{1}{c|}{N/A}             & \multicolumn{1}{c|}{N/A}            &  \multicolumn{1}{c|}{N/A}                             & \multicolumn{1}{l|}{$48.4 \pm 0.4$}             & \multicolumn{1}{l|}{$50.1 \pm 0.9$}            &        \multicolumn{1}{l}{$0.014 \pm 0.00$}                        \\ \hline
Ground-truth             & \multicolumn{1}{l|}{$62.2 \pm 0.5$}             & \multicolumn{1}{l|}{$22.4 \pm 1.1$}            &       \multicolumn{1}{l|}{$0.022 \pm 0.01$}                          & \multicolumn{1}{l|}{$53.9 \pm 0.4$}             & \multicolumn{1}{l|}{$55.5 \pm 1.0$}            &        \multicolumn{1}{l}{$0.014 \pm 0.00$}                        \\ \hline
HyperBO+                 & \multicolumn{1}{l|}{$62.2 \pm 0.5$}             & \multicolumn{1}{l|}{$22.8 \pm 1.0$}            &              \multicolumn{1}{l|}{$0.032 \pm 0.01$}                   & \multicolumn{1}{l|}{$54.2 \pm 0.4$}             & \multicolumn{1}{l|}{$56.1 \pm 1.0$}            &         \multicolumn{1}{l}{$0.015 \pm 0.00$}                       \\ \hline
$^{\mathsf{x}}$HyperBO+                 & \multicolumn{1}{l|}{$61.9 \pm 0.2$}             & \multicolumn{1}{l|}{$22.2 \pm 0.7$}            &              \multicolumn{1}{l|}{$0.030 \pm 0.00$}                   & \multicolumn{1}{l|}{$53.4 \pm 0.3$}             & \multicolumn{1}{l|}{$55.1 \pm 0.6$}            &         \multicolumn{1}{l}{$0.013 \pm 0.00$}                       \\ \hline
\end{tabular}
\end{adjustbox}
\end{center}
\end{table*}

\begin{table}[!htpb]
\caption{NLL evaluations of the training and testing data on different models and normalized simple regret of running BO with each method (PI acquisition function) for the HPO-B Super-dataset. ($\mathrm{mean} \pm \mathrm{std}$)}
\label{tab:nll-regret-hpob}
\begin{center}
\begin{adjustbox}{width=0.6\textwidth}
\begin{tabular}{c|l|l|l}
\hline
Methods         & \multicolumn{1}{c|}{Train NLL} & \multicolumn{1}{c|}{Test NLL} & \multicolumn{1}{c}{BO Regret} \\ \hline
Non-informative &   \multicolumn{1}{l|}{$90.5 \pm 0.0$}            &  \multicolumn{1}{l|}{$96.7 \pm 0.4$}              &  \multicolumn{1}{l}{$0.088 \pm 0.01$}                              \\ \hline
Hand-specified  &\multicolumn{1}{l|}{$-23.9 \pm 0.0$} & \multicolumn{1}{l|}{$-26.3 \pm 0.1$} & \multicolumn{1}{l}{$0.091 \pm 0.01$}  \\ \hline
HyperBO         &  \multicolumn{1}{l|}{$-197.1 \pm 0.8$} & \multicolumn{1}{l|}{$-180.7 \pm 2.5$}                          &  \multicolumn{1}{l}{$0.044 \pm 0.01$}                              \\ \hline
HyperBO+        &   \multicolumn{1}{l|}{$-261.4 \pm 0.4$} & \multicolumn{1}{l|}{$-272.9 \pm 1.9$}                     &     \multicolumn{1}{l}{$0.038 \pm 0.00$}                           \\ \hline
$^{\mathsf{z}}$HyperBO+        &   \multicolumn{1}{l|}{$-256.0 \pm 0.7$} & \multicolumn{1}{l|}{$-268.3 \pm 2.1$}                        &    \multicolumn{1}{l}{$0.041 \pm 0.00$}                            \\ \hline
$^{\mathsf{x}}$HyperBO+        &   \multicolumn{1}{l|}{$-263.2 \pm 0.5$} & \multicolumn{1}{l|}{$-273.5 \pm 2.1$}                        &    \multicolumn{1}{l}{$0.041 \pm 0.01$}                            \\ \hline
\end{tabular}
\end{adjustbox}
\end{center}
\end{table}


Fig.~\ref{fig:bo-regrets-eipi-synthetic} shows the BO performances of compared methods in both Setup A and Setup B for the Synthetic Super-dataset, and Fig.~\ref{fig:bo-regrets-eipi-hpob} shows the results for the HPO-B Super-dataset. 
The budget for BO is 50, and there are 5 initial observations that are randomly sampled for each of the 5 random seeds. The acquisition function used for all GP-based methods is \textit{Probability of Improvement} (PI)~\citep{kushner1964} with target value $\max(y_t)+0.1$. As shown by~\cite{wang2017maxvalue}, PI can obtain high BO performance by setting good target values. Results with other types of acquisition functions are included in the supplement.

In Setup A for the Synthetic Super-dataset (Fig.~\ref{fig:bo-regrets-eipi-synthetic}(a)), HyperBO+ outperforms the non-informative hierarchical GP, the hand-specified hierarchical GP, and the random baseline, demonstrating its ability to learn a universal prior from training datasets and generalizing to tasks in new search spaces that are unseen during pre-training. HyperBO+ achieves good BO performance similar to the ground-truth model. This shows that the two-step pre-training method is very effective in learning the prior parameters. 

In Setup B for the Synthetic Super-dataset (Fig.~\ref{fig:bo-regrets-eipi-synthetic}(b)), HyperBO+ again achieves lower regrets than the non-informative hierarchical GP, the hand-specified hierarchical GP, and the random baseline. HyperBO+ and HyperBO achieve BO performances similar to using the ground-truth prior. $^{\mathsf{x}}$HyperBO+ achieves a slightly lower regret than the standard HyperBO+ in both setups.

For the HPO-B Super-dataset (Fig.~\ref{fig:bo-regrets-eipi-hpob}), HyperBO+ achieves lower regrets than the non-informative hierarchical GP, the hand-specified hierarchical GP, and the random baseline. Notably, HyperBO+ outperforms HyperBO on HPO-B, which is rather surprising given that HyperBO specifically pre-trains a GP that targets a single search space. In contrast, HyperBO+ learns one universal prior on training sub-datasets from all search spaces, and the learned universal prior would have to automatically capture the shared characteristics of testing functions in different search spaces. 

We conjecture that the performance difference is because HyperBO overfits or fits an incorrect GP for some search spaces which damages its performance on the testing sub-datasets from those search space. Meanwhile, HyperBO+ uses a single learned prior for BO in all search spaces, which to some extent regularizes an incorrect GP learned on a particular search space with GPs learned from other search spaces. Fig.~\ref{fig:nll-per-search-space-hpob} shows that HyperBO can obtain high test NLLs despite having much lower train NLLs in some search spaces of HPO-B.

Notably, $^{\mathsf{z}}$HyperBO+ outperforms the baselines with only slightly worse performance than the standard HyperBO+ even though it is not trained on the training sub-datasets of the search space it is tested in. The hyperparameters in tuning tasks of HPO-B can be entirely irrelevant, e.g. hyperparameters for SVM and Random Forest. As a result, the training data for $^{\mathsf{z}}$HyperBO+ may not even be equipped to inform new tasks for testing. The good performance of  $^{\mathsf{z}}$HyperBO+ further demonstrates the capability of HyperBO+ to generalize across search spaces in real-world problems. 

Interestingly, $^{\mathsf{x}}$HyperBO+ achieves a similar and slightly worse performance than the standard HyperBO+. Recall that  $^{\mathsf{x}}$HyperBO+ does not use the MLE step to estimate parameterized priors, and instead, just use the uniform distribution over the parameters of mean and kernel functions.  Fig.~\ref{fig:lengthscale-dist-hpob} shows the histogram of the length-scale values learned from the pre-training phase, and the probability density of the universal Gamma prior on length-scales learned by HyperBO+ in Line~\ref{alg:mle} of Alg.~\ref{alg:hyperbo}. Despite the fact that our universal Gamma prior assigns low probability at length-scale=$9$, which is inconsistent with the histogram, using the pre-trained Gamma prior as part of the standard HyperBO+ still performs very well or even slightly better than $^{\mathsf{x}}$HyperBO+. The results here also demonstrate that some training data can have negative impacts on testing tasks, but HyperBO+ is able to mitigate the negative transfer learning issues.

\subsection{Evaluations of NLLs}
\label{subsec:evaluation-of-nlls}

Tab.~\ref{tab:nll-regret-synthetic} shows the NLL evaluations of training/testing data on the compared models together with the BO regret in both Setup A and Setup B for the Synthetic Super-dataset, while Tab.~\ref{tab:nll-regret-synthetic} shows the results for the HPO-B super-dataset.

The results show that HyperBO+ achieves a much lower NLL than the hand-specified hierarchical GP on both super-datasets, and achieves a similar NLL as the ground-truth hierarchical GP in both setups on the Synthetic Super-dataset. This further demonstrates the effectiveness of the two-step pre-training of HyperBO+. The two variants of HyperBO+, $^{\mathsf{z}}$HyperBO+ and $^{\mathsf{x}}$HyperBO+, both achieve a similar NLL to the standard HyperBO+.

In Setup B, the original HyperBO gets a lower NLL than HyperBO+ and the ground-truth hierarchical GP on the Synthetic Super-dataset, which makes sense as HyperBO specifically pre-trains a GP that targets a single search space. However, HyperBO has higher NLLs than HyperBO+ on the HPO-B super-dataset. This is in line with our observation of BO performances on HPO-B. The reason for this could to be similar to what we conjectured in \S\ref{subsec:bo-results} - that is, HyperBO overfits in some of the search spaces of HPO-B, while HyperBO+ is capable of regularizing incorrect GP parameters learned in these parameters with GPs learned in other search spaces because it uses the same universal prior for all search spaces.

\section{Conclusions}
\label{sec:conclusions}

We present HyperBO+, the first GP-based transfer learning BO method that works on universal search spaces. The key idea is to pre-train a universal hierarchical GP prior on training data partitioned over search spaces and functions within each search space.
We demonstrate the appealing asymptotic properties of our pre-training method and show HyperBO+ outperforms competitive baselines through experiments on challenging hyperparameter tuning tasks.


\subsection*{Acknowledgments}
We thank Jasper Snoek and Eytan Bakshy for helpful conversations and/or feedback. Our work also benefited from Microsoft Azure credits provided by the Harvard Data Science Initiative, as well as Google Cloud Platform Credit Award provided by Google.

\bibliography{ref}

\onecolumn
\appendix

\section{Discussions}
\paragraph{Limitations and future work.} HyperBO+ learns a prior distribution on Gaussian processes (GPs) from existing data to improve the performance of Bayesian optimization (BO) on new task. However, there are other components that can also be meta-learned, such as acquisition functions~\citep{volpp2020meta}, to maximize the effectiveness of the BO system. One direction of future work is jointly pre-train all components of BO to allow more flexibility and further improve the performance. In addition, HyperBO+ currently assumes a stationary kernel function and constant mean function when fitting the GP parameters. Possible future work includes relaxing this assumption and incorporating architecture search for the kernel function and mean function to enrich the space of hierarchical GP priors.

\paragraph{Societal impact of this work.}
Hyperparameter tuning for machine learning (ML) models, especially deep learning models, can be very costly if we repeatedly evaluate a large number of hyperparameters. Each single evaluation of a hyperparameter value requires training and testing a new instantiation of the model. Our framework HyperBO+, with its superior BO performance discussed in \S\ref{subsec:bo-results}, can help to reduce the number of evaluations needed for hyperparameter tuning tasks and thus reduce their computational cost and carbon footprint potentially by a large margin.

\section{More details of theoretical analysis}
\label{app:theory}
\paragraph{Proof of Lemma~\ref{prop:increasing}}
\proof
Consider functions $f_{ij}$ whose observations $D_{ij} = \{(x_{k}^{ij},\, y_{k}^{ij})\}_{k=1}^{L_{ij}}$ is in dataset $D_i$, where $j\in \{1, \dots, M_i\}$.

Since the domain $\mathcal X_i$ of $f_{ij}$ is compact, and we define $Q' = \sup_{x, x'\in \mathcal X_i} \|x' - x\| + 1\ll \infty$.  Since the stationary covariance function only concerns the relative location between two points, we can shift all inputs jointly while the NLL $L(\theta \mid \{D_{ij}\})$ stays the same. More formally, let $\bar D_j = \{(x_{k}^{ij} + 1_{d_i} (Q + Q')j,\,  y_{k}^{ij})\}_{k=1}^{L_{ij}}$ for some $Q > 0$, where $d_i$ is the input dimension and $1_{d_i}$ is a $d_i-$dimensional vector filled with ones. Because the distance between any pairs of inputs stays the same, we have $L(\theta \mid \{D_{ij}\}) = L(\theta \mid  \{\bar D_j\}) = -\log p( D_{ij} \mid \theta)$.

We define an augmented sub-dataset $\bar D = \bar D_{1}\cup \cdots\cup \bar D_{M_i}$. 

As $Q \rightarrow \infty$, we can show that the sum of NLLs over a set of sub-datasets and the NLL of the single augmented sub-dataset are equivalent. Without loss of generality, consider the Mat\'ern covariance function with smoothness term $\nu = 1/2$ and variance $\sigma^2$, length-scale $\rho$: $C_{1/2}(d) = \sigma^2 \exp (-\frac{d}{\rho})$. Since the covariance exponentially decays with the distance $d$ between points, for any $\epsilon > 0$, we can find a $Q$ such that $C(d) < \epsilon$ for $d > Q$. 

For the augmented sub-dataset $\bar D = \bar D_{1}\cup \cdots\cup \bar D_{M_i}$, we have $\|x' - x\| > Q$ for any $x\in \bar D_j$ and $x' \in \bar D_{j'}$, $j \neq j'$. As $\epsilon \rightarrow 0$, the covariance matrix for this augmented sub-dataset becomes block diagonal. As a result, we can express its NLL as a sum of NLL for $D_{i1}, \dots, D_{iM_i}$ respectively, i.e. $L(\theta \mid \{\bar D\}) = -\sum_{j=1}^{M_i} \log p(\bar D_{j} \mid \theta)$. This gives us the equivalence in Eq.~\ref{eq:prop}. 

Note that the distance between inputs in $\bar D$ will always be at least $\min_j\min_{k \neq k'} \|x^{ij}_k - x^{ij}_{k'}\| > \delta$, since any two datapoints in each sub-dataset $D_{ij}$ are at least $\delta$ apart by construction in problem setup.
As $M_i \rightarrow \infty$, the augmented sub-dataset $\bar D$ is unbounded and satisfies the increasing domain characterization in Eq.~\ref{eq:increasing}.
\endproof%

Another interesting related work is from \cite{karvonen2022maximum}, where the authors pointed out that the MLE for the length-scale parameter is indeed ill-posed when the data is noiseless and can be unstable to small perturbations. However, they found that regularization with small additive Gaussian noise
does guarantee well-posedness and this is equivalent to data corrupted by additive Gaussian noise. This is identical to our setting with the synthetic data and we found that in practice an accurate estimation for noise allows reliable inference for covariance parameters.

\paragraph{Proof of Theorem~\ref{theorem3}}

\proof
By Lemma~\ref{prop:increasing}, our equivalence setup satisfies the conditions in Theorem 1 from \cite{bachoc2021asymptotic}. Therefore, the conclusion holds that $\hat{\theta}_{ML} \rightarrow^p \theta_0$. \qed

\paragraph{Proof of Theorem~\ref{theorem4}}

\proof
The asymptotic properties for the second training step follows from standard statistics literature \citep{wackerly2014mathematical} since we assume each datasets' covariance parameters are drawn \textit{i.i.d.} from some prior distribution. \endproof

\section{More details of the experiment setups}
\label{app:exp_setups}
Our code for the experiments is built upon the codebase of HyperBO~\citep{wang2022pre}, and is available at \url{https://github.com/Evensgn/hyperbo}.

The Synthetic Super-dataset includes 20 datasets (search spaces) with 10 sub-datasets in each dataset. Each sub-dataset includes noisy observations at 300 input locations in its respective search space. The dimensions of the search spaces are randomly sampled between 2 and 5. As all the datasets are \textit{i.i.d.} samples, we use the first $16$ datasets to be training datasets and the remaining $4$ datasets to be testing datasets in Setup A. Similarly, we use the first $8$ sub-datasets of each dataset to form the training data and use the remaining $2$ sub-datasets of each dataset to form the testing data in Setup B.

HPO-B Super-dataset includes 16 datasets (search spaces), while each dataset contains different numbers of sub-datasets and has its pre-specified train/test split of sub-datasets. We use this pre-specified train/test split for the experiments on HPO-B.

In the experiments, all GP-based methods use a constant mean function and a Mat\'ern kernel with smoothness parameter $\nu=3/2$. In our setting, each dataset $D_i$ ($i = 1, \dots, N$) corresponds to $\mathcal{GP}_i$ parameterized by $\theta_i$, which includes the following parameters: constant mean (the value of the constant mean function), length-scale (which has the same number of dimensions as the search space), signal variance. In addition, $D_i$ is also parameterized by the noise variance $\sigma_i$. 

For the prior distribution types of each of these parameters, we use a normal distribution for constant mean and use a Gamma distribution for each of the remaining parameters. A normal distribution is parameterized by $\mu$ (mean) and $\sigma$ (standard deviation), while a Gamma distribution is parameterized by $\alpha$ (shape) and $\beta$ (rate).

The specific distribution parameters used to generate the Synthetic Super-dataset are as follows: constant mean is sampled from $\mathrm{Normal}(\mu=1, \sigma=1)$, each dimension of length-scale is sampled from $\mathrm{Gamma}(\alpha=10, \beta=30)$, signal variance is sampled from $\mathrm{Gamma}(\alpha=1, \beta=1)$, and noise variance is sampled from $\mathrm{Gamma}(\alpha=10, \beta=100000)$. 

For the non-informative hierarchical GP, the specific parameters are as follows: constant mean is sampled from $\mathrm{Uniform}(low=-100, high=100)$, each dimension of length-scale is sampled from $\mathrm{Uniform}(low=0.001, high=10)$, signal variance is sampled from $\mathrm{Uniform}(low=0.000001, high=100)$, and noise variance is sampled from $\mathrm{Uniform}(low=0.00000001, high=100)$.

For the hand-specified hierarchical GP, the specific parameters are as follows: constant mean is sampled from $\mathrm{Normal}(\mu=0, \sigma=1)$, each dimension of length-scale is sampled from $\mathrm{Gamma}(\alpha=1, \beta=10)$, signal variance is sampled from $\mathrm{Gamma}(\alpha=1, \beta=5)$, and noise variance is sampled from $\mathrm{Gamma}(\alpha=10, \beta=100)$.

We fit the GP parameters in each search space by minimizing Eq.~\ref{eq:nll} using L-BFGS for the Synthetic Super-dataset, and using the Adam optimizer for HPO-B. For L-BFGS on the Synthetic Super-dataset, the number of iterations is 500, and all observations from a sub-dataset is used at each iteration of L-BFGS. For Adam optimizer on HPO-B Super-dataset, the number of iterations is 10000, and each sub-dataset is sub-sampled to 50 observations at each iteration of Adam optimizer. The learning rate of Adam optimizer is 0.001. The reason for using Adam in HPO-B experiments is that the sub-datasets in HPO-B can be too large to be handled by L-BFGS. Furthermore, L-BFGS was recommended by~\cite{wang2022pre} as the standard objective optimization method for HyperBO, while Adam was recommended by \cite{wistuba2021few} that applied the FSBO method on HPO-B.

When testing the BO performances of compared methods, the BO budget is 50, and 5 random observations are made prior to BO for initialization.  For hierarchical GP-based methods, 100 samples of GP parameters are used to compute the acquisition function value according to Eq.~\ref{equ:ac_fun} at each BO iteration. All methods are tested on 5 random seeds, while the set of initial observations are the same for all methods given the same random seed. Each sub-dataset of HPO-B are sub-sampled to 1000 observations for BO testing, while no sub-sampling is needed for the Synthetic Super-dataset since each sub-dataset contains 300 observations.

For evaluation of NLL values, each sub-dataset is sub-sampled to 100 observations due to the time complexity of computing NLL, and we report the average NLL across 10 random seeds. The sub-sampling is the same for all compared methods given the same random seed. For hierarchical GP-based methods, 500 samples of GP parameters are used to compute the NLL value according to Eq.~\ref{equ:estimate-nll}.

\section{Additional experiment results}
\label{app:more_exp}
In this section, we present additional empirical evidences on the asymptotic behaviors of our pre-training method, as well as analyses on the performance on Bayesian optimization.
\subsection{Asymptotic behavior of fitting a single GP}
\begin{figure}
    \centering
    \begin{subfigure}{0.42\textwidth}
        \includegraphics[width=\linewidth]{images/lengthscale_n_sub_datasets_25_observations.pdf}
        \caption{Length-scale}
        \label{fig:one-gp-asymptotic-lengthscale-copy}
    \end{subfigure}
    \vfill
    \begin{subfigure}{0.42\textwidth}
        \includegraphics[width=\linewidth]{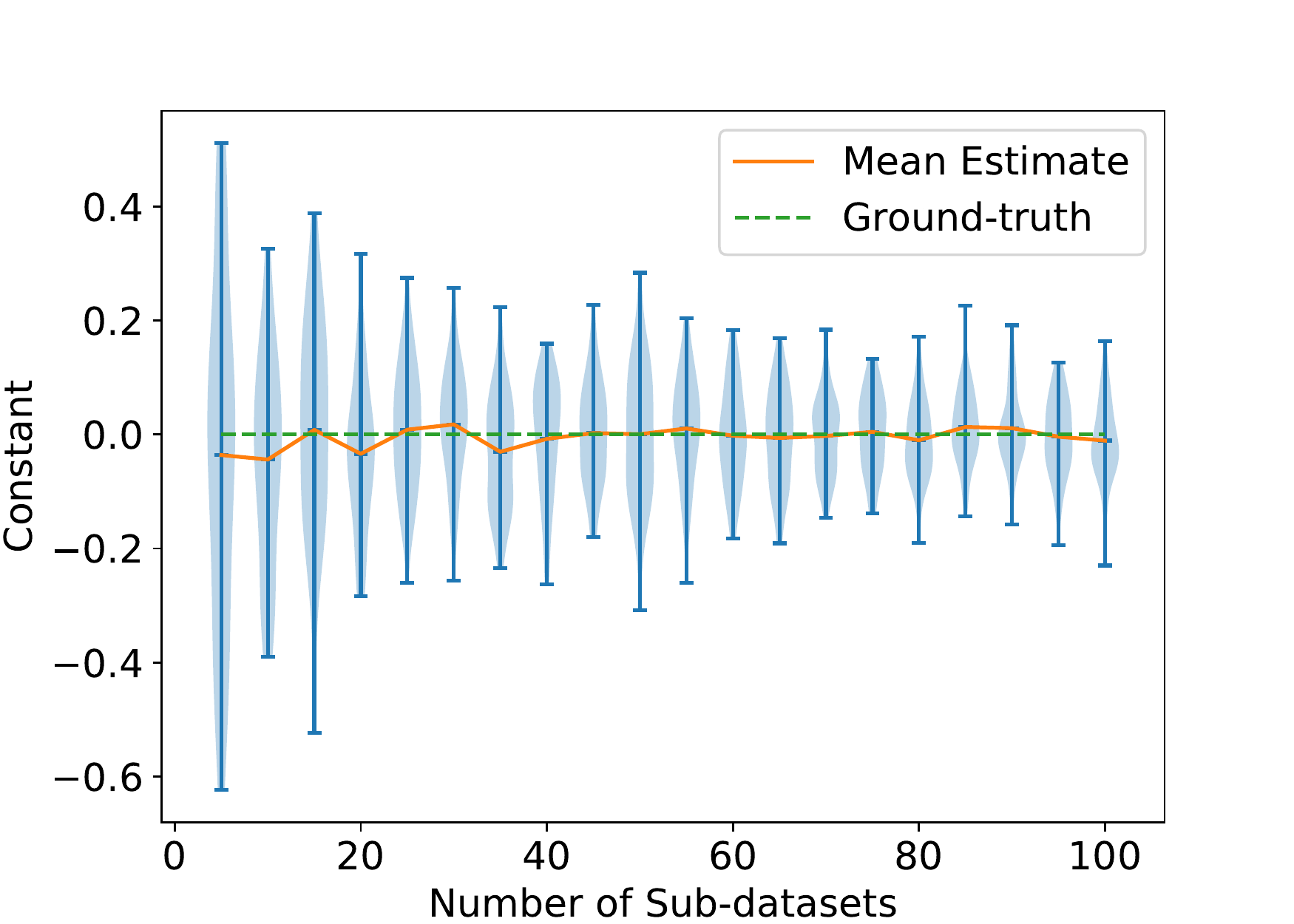}
        \caption{Constant Mean}
        \label{fig:one-gp-asymptotic-constant}
    \end{subfigure}
    \vfill
    \begin{subfigure}{0.42\textwidth}
        \includegraphics[width=\linewidth]{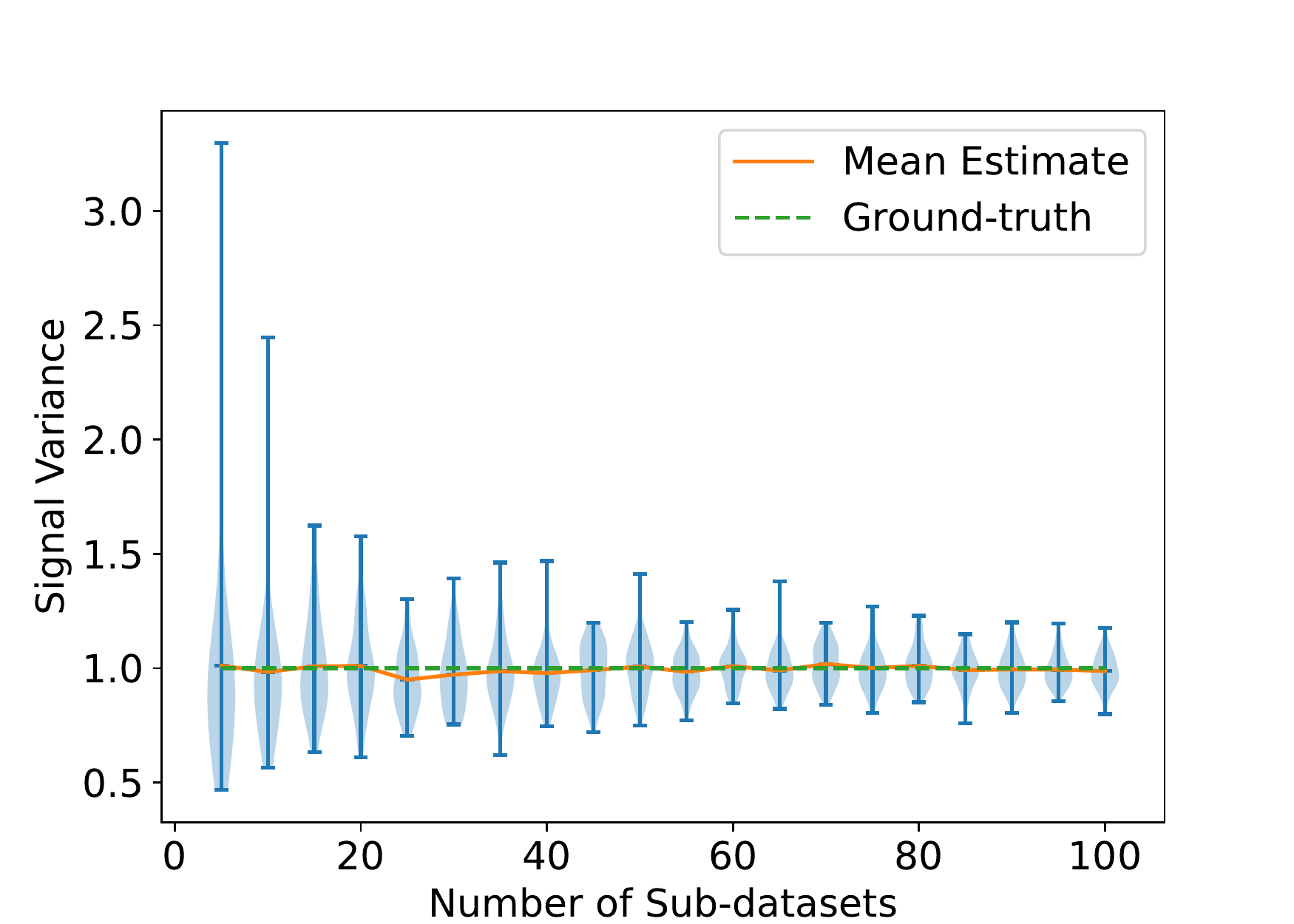}
        \caption{Signal Variance}
        \label{fig:one-gp-asymptotic-signal-variance}
    \end{subfigure}
    \vfill
    \begin{subfigure}{0.42\textwidth}
        \includegraphics[width=\linewidth]{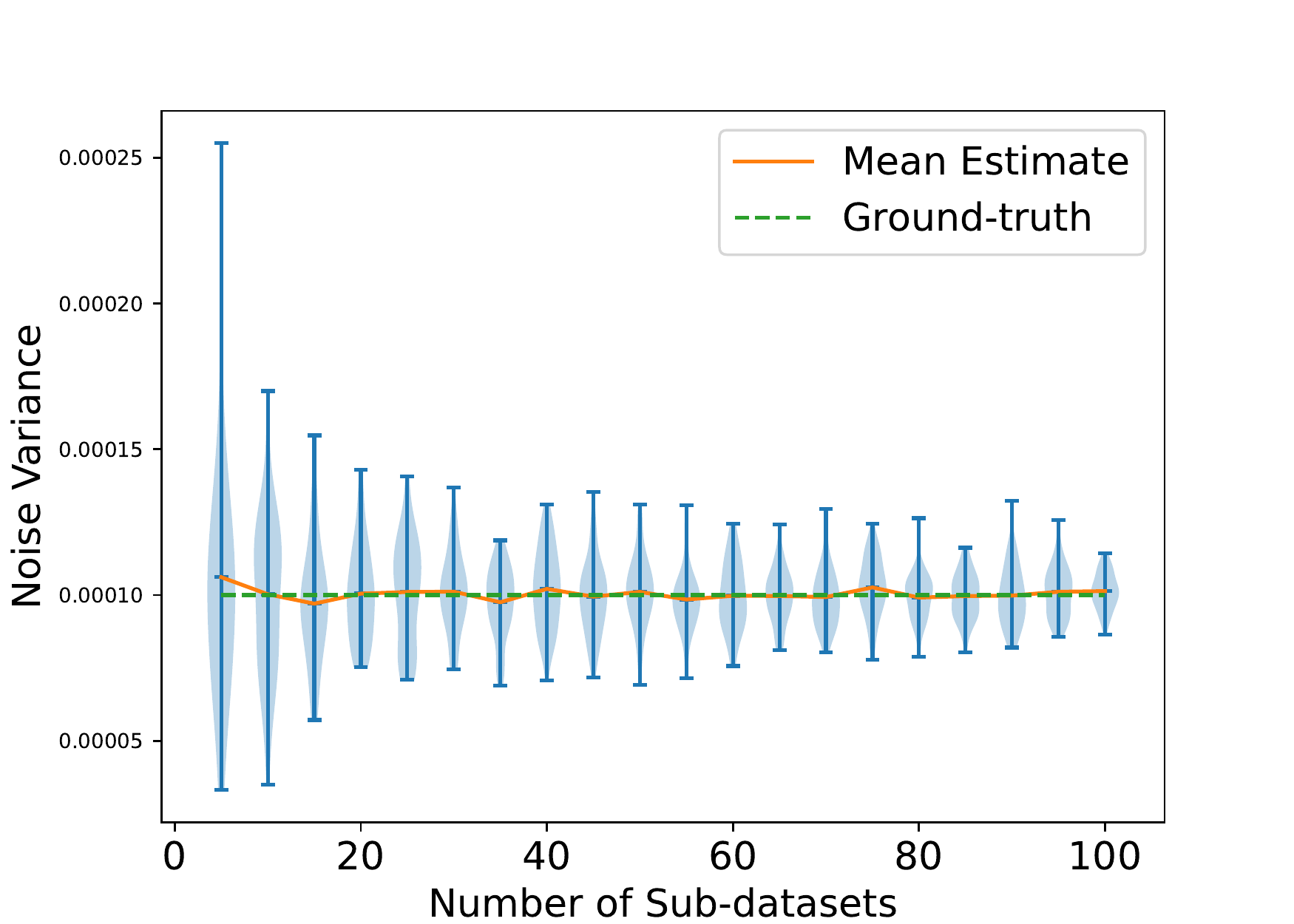}
        \caption{Noise Variance}
        \label{fig:one-gp-asymptotic-noise-variance}
    \end{subfigure}
    \caption{Estimated length-scale, constant mean, signal variance and noise variance parameters of a 1-dimensional GP as the number of sub-datasets increases. Each sub-dataset contains 25 observations. The violin plot shows the distribution of estimates for each number of sub-datasets over 50 random seeds. The variance of estimated GP parameters decreases as the number of sub-datasets increases.}
    \label{fig:one-gp-asymptotic-constant-sigvar-noisevar}
\end{figure}

 Fig.~\ref{fig:one-gp-asymptotic-constant-sigvar-noisevar} demonstrates the asymptotic behavior empirically for fitting the parameters of a single GP using an increasing number of sub-datasets with simulations on synthetic data generated with a 1-dimensional GP. The ground-truth GP parameters are shown in the figure. The variance of estimated GP parameters decreases as the number of sub-datasets increases.

\begin{figure}
    \centering
    \includegraphics[width=\linewidth]{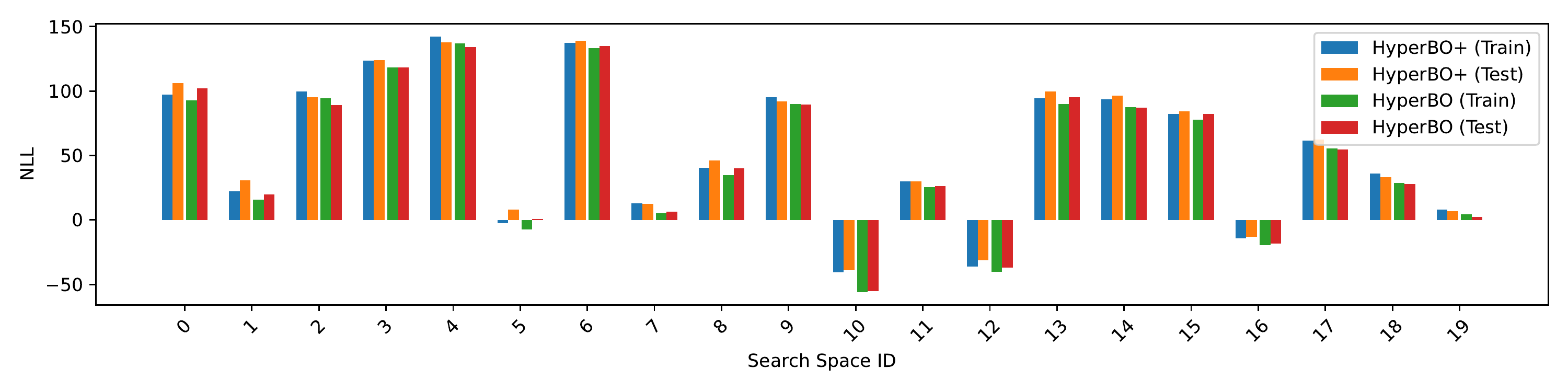}
    \caption{NLL evaluations of training/testing data on HyperBO+ and HyperBO in Setup B on the Synthetic Super-dataset.}
    \label{fig:nll-per-search-space-synthetic}
\end{figure}

\subsection{NLL evaluations per search space in Setup B} 

Fig.~\ref{fig:nll-per-search-space-hpob} shows the NLL evaluations of training/testing data on HyperBO+ and HyperBO for the HPO-B Super-dataset. HyperBO obtained high test NLLs despite having much lower train NLLs in some search spaces (such as 5527, 5891, and 6794). This observation supports our conjecture in \S\ref{subsec:bo-results} that HyperBO overfits in some search spaces of HPO-B. Fig.~\ref{fig:nll-per-search-space-synthetic} shows the NLL evaluations of training/testing data on HyperBO+ and HyperBO in Setup B on the Synthetic Super-dataset. We can see that the overfitting does not occure on the Synthetic Super-dataset as HyperBO+ has slightly higher NLL values than HyperBO. The reason for this difference across the two super-datasets might be that the Synthetic Super-datasets are generated so that all sub-datasets in a search space are sampled from the same GP, while the HPO-B Super-dataset does not guarantee this property as its data are collected from real-world hyperparameter tuning tasks.

\begin{figure}
    \vspace{-10pt}
    \centering
    \begin{subfigure}{0.40\textwidth}
        \includegraphics[width=\linewidth]{images/mle_lengthscale_n_train_datasets.pdf}
        \caption{Length-scale - $\alpha$ of Gamma Distribution}
        \label{fig:two-step-asymptotic-lengthscale-synthetic-copy}
    \end{subfigure}
    \begin{subfigure}{0.40\textwidth}
        \includegraphics[width=\linewidth]{images/mle_distribution_lengthscale_n_train_datasets.pdf}
        \caption{Length-scale - Gamma Distribution}
        \label{fig:two-step-asymptotic-lengthscale-synthetic-distribution}
    \end{subfigure}
    \vfill
    \begin{subfigure}{0.40\textwidth}
        \includegraphics[width=\linewidth]{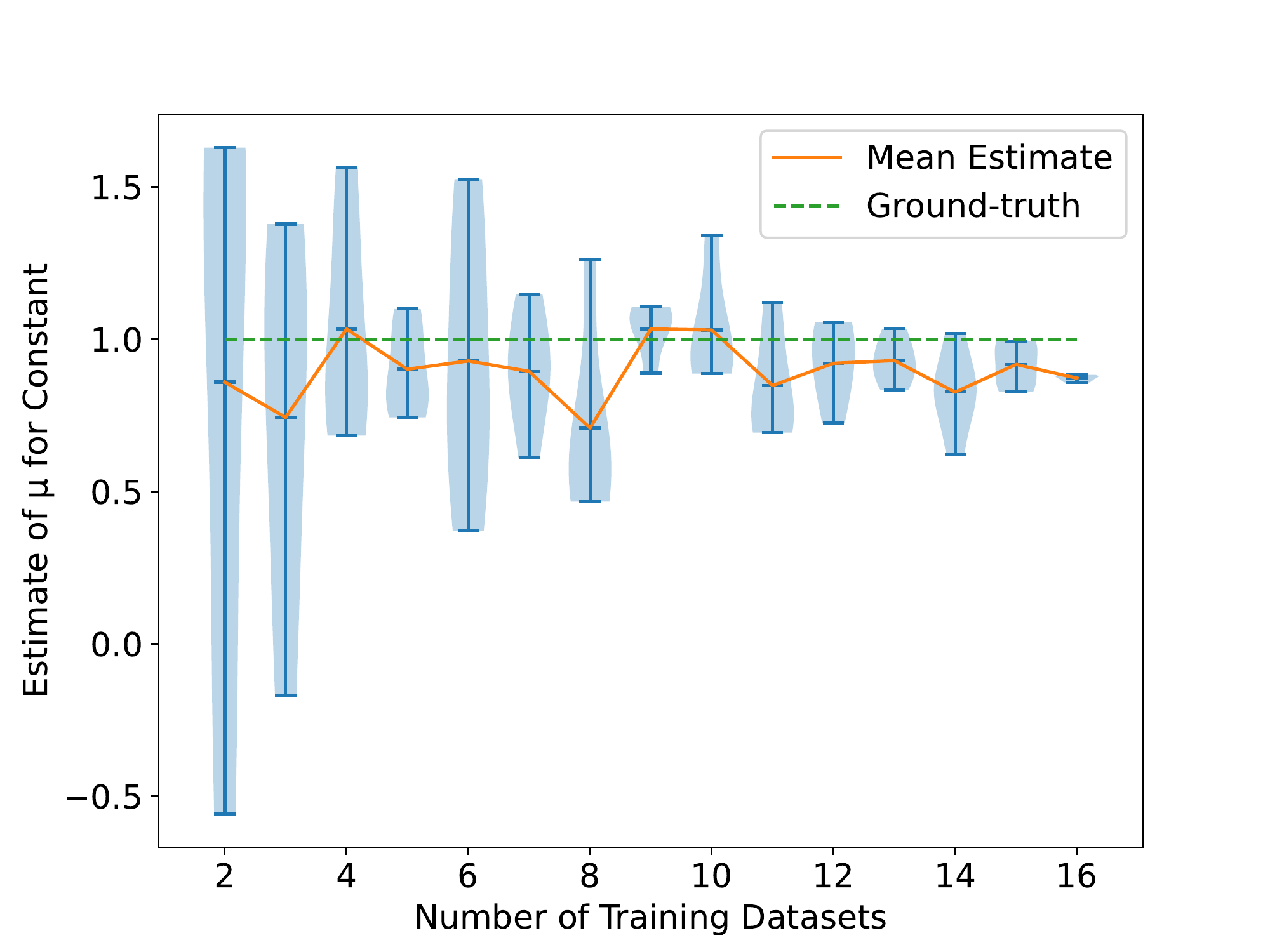}
        \caption{Constant Mean - $\mu$ of Normal Distribution}
        \label{fig:two-step-asymptotic-constant-synthetic}
    \end{subfigure}
    \begin{subfigure}{0.40\textwidth}
        \includegraphics[width=\linewidth]{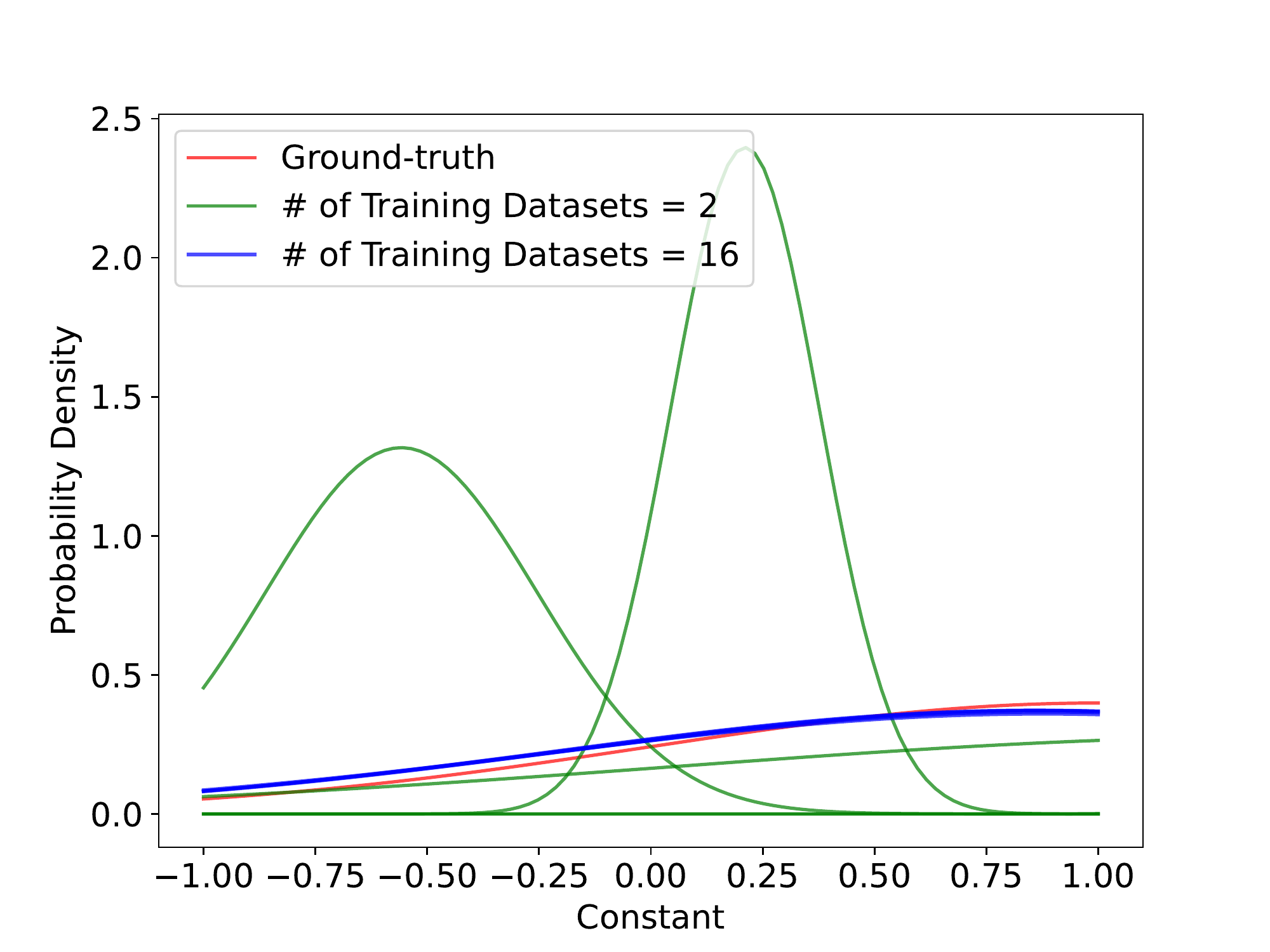}
        \caption{Constant Mean - Normal Distribution}
        \label{fig:two-step-asymptotic-constant-synthetic-distribution}
    \end{subfigure}
    \vfill
    \begin{subfigure}{0.40\textwidth}
        \includegraphics[width=\linewidth]{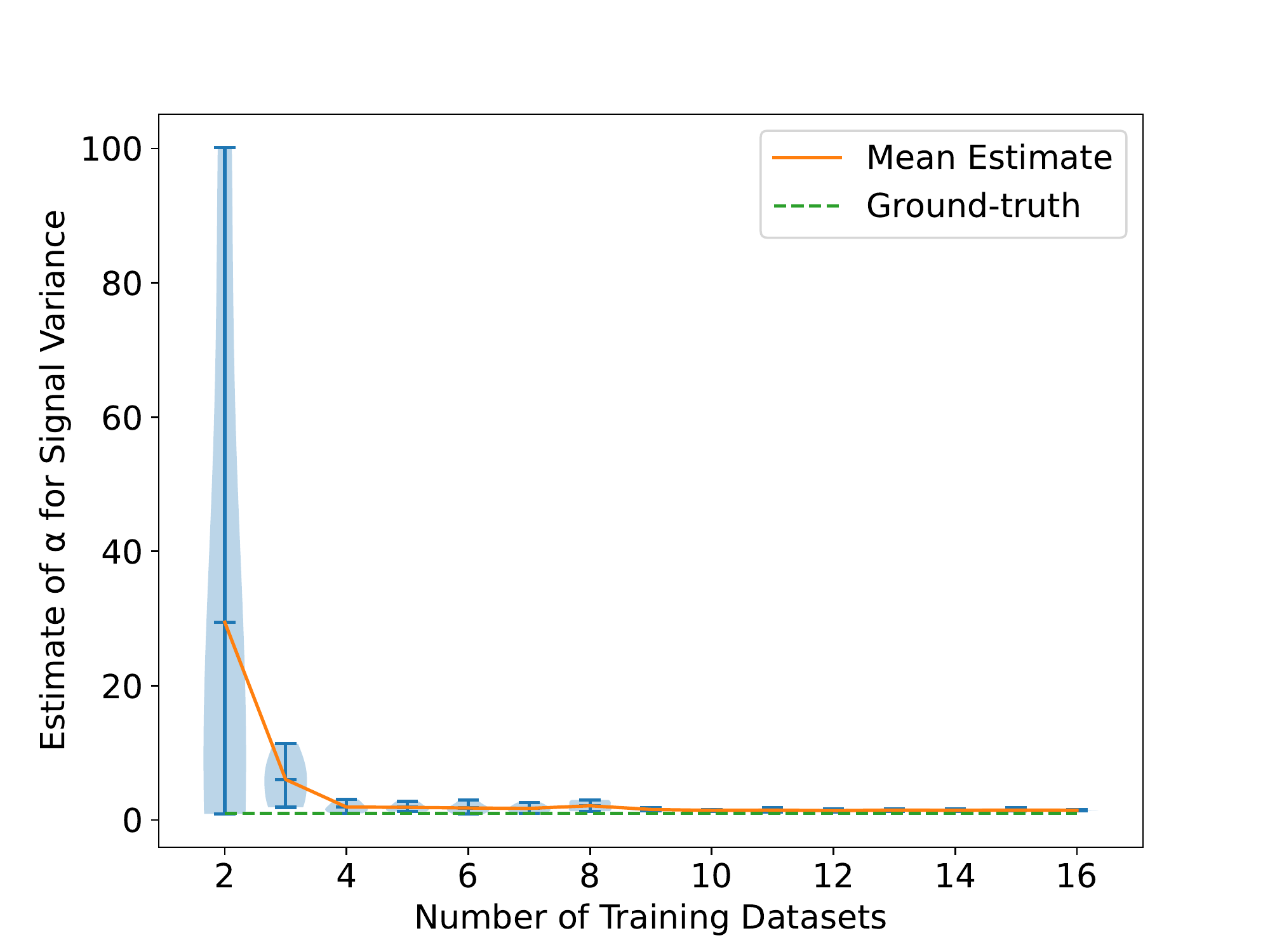}
        \caption{Signal Variance - $\alpha$ of Gamma Distribution}
        \label{fig:two-step-asymptotic-signal-variance-synthetic}
    \end{subfigure}
    \begin{subfigure}{0.40\textwidth}
        \includegraphics[width=\linewidth]{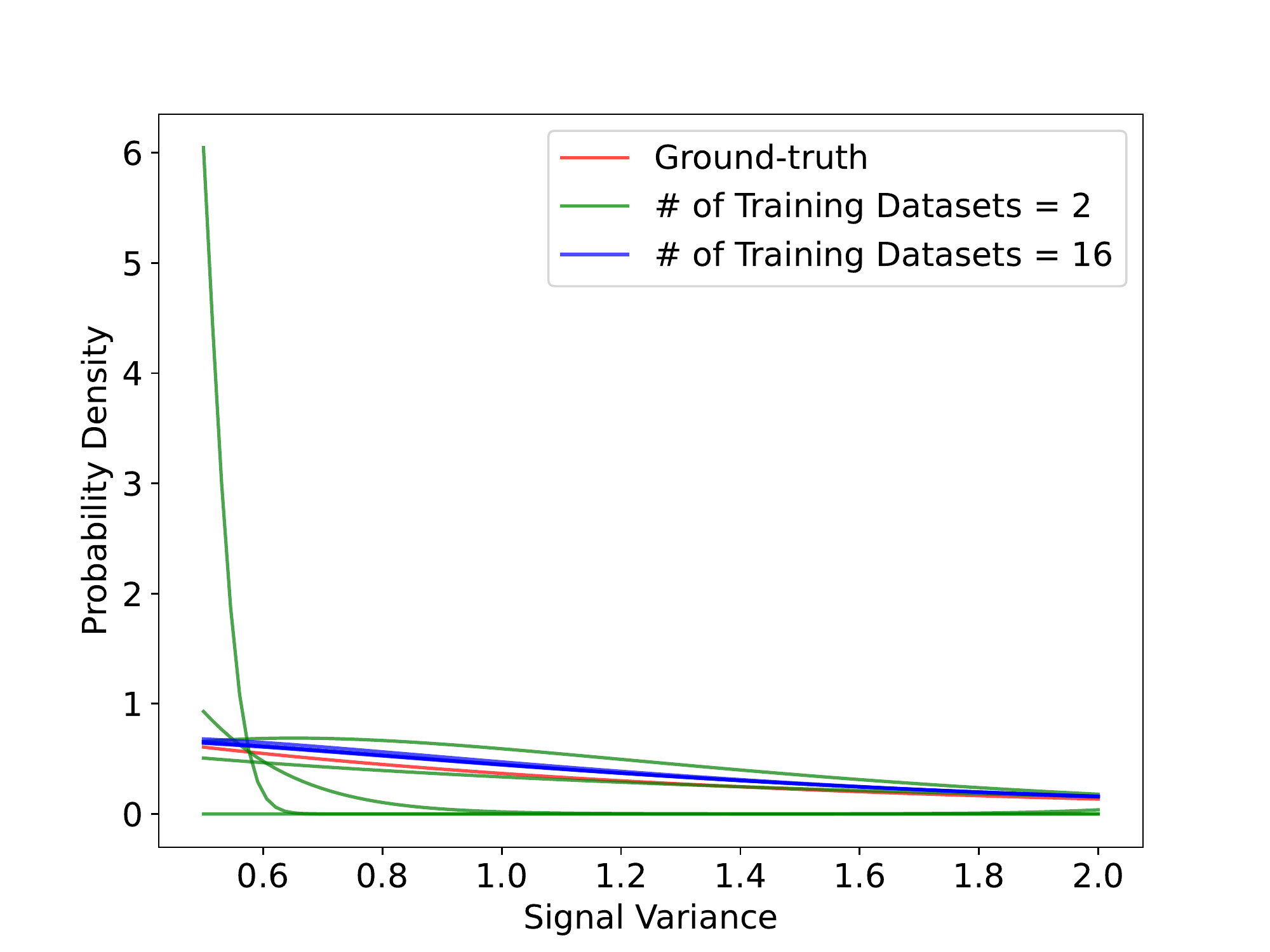}
        \caption{Signal Variance - Gamma Distribution}
        \label{fig:two-step-asymptotic-signal-variance-synthetic-distribution}
    \end{subfigure}
    \vfill
    \begin{subfigure}{0.40\textwidth}
        \includegraphics[width=\linewidth]{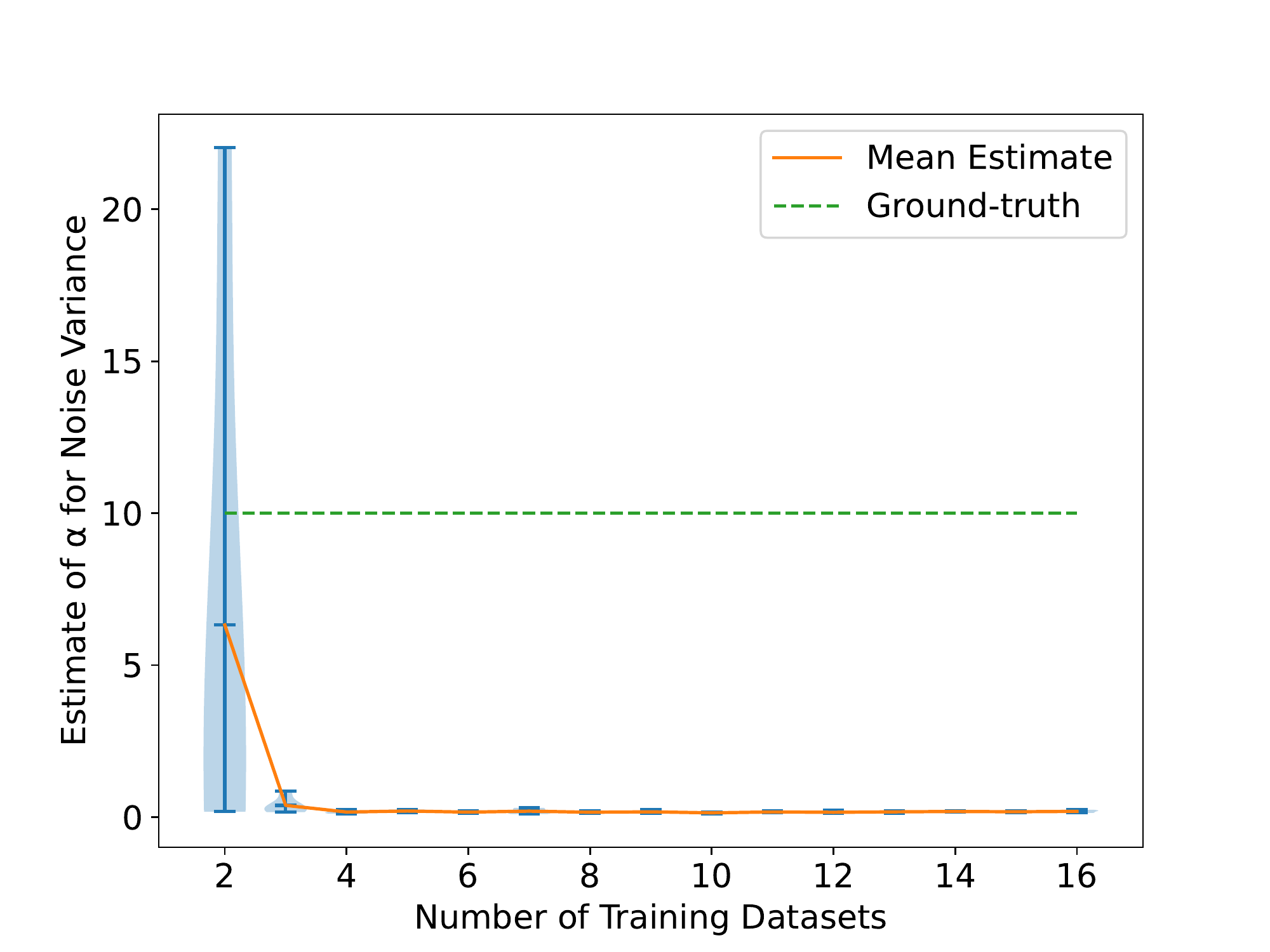}
        \caption{Noise Variance - $\alpha$ of Gamma Distribution}
        \label{fig:two-step-asymptotic-noise-variance-synthetic}
    \end{subfigure}
    \begin{subfigure}{0.40\textwidth}
        \includegraphics[width=\linewidth]{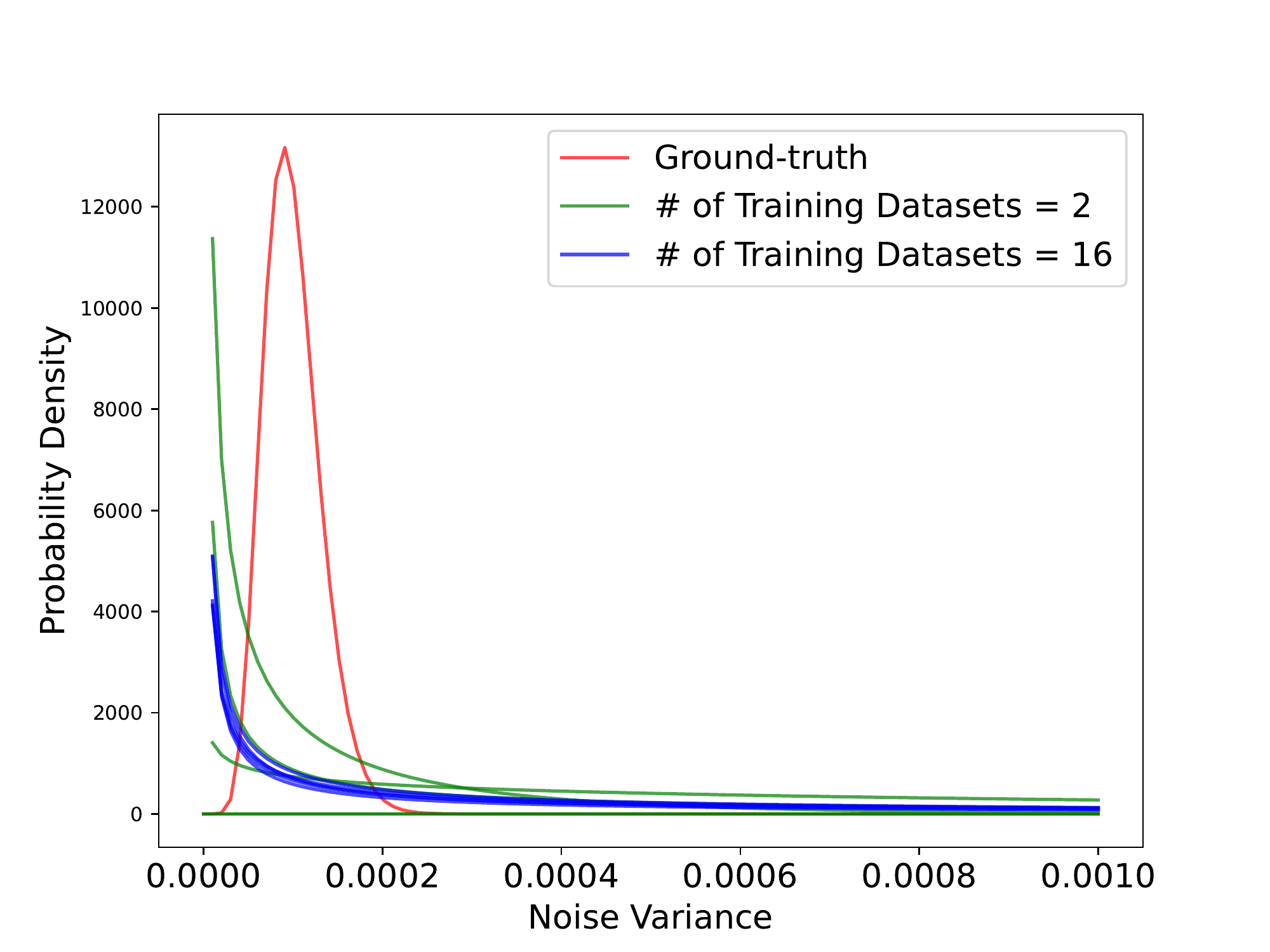}
        \caption{Noise Variance - Gamma Distribution}
        \label{fig:two-step-asymptotic-noise-variance-synthetic-distribution}
    \end{subfigure}
    \caption{Left: Estimated shape parameter $\hat{\alpha}$ (y-axis) of Gamma distribution prior for the length-scale, signal variance and noise variance parameters of a kernel function and estimated mean parameter $\hat{\mu}$ (y-axis) of the normal distribution prior that models the constant mean GP parameter w.r.t. the number of training datasets (x-axis). We show the mean and violin plots for $\hat{\alpha}$ over 5 random seeds. Right: The pre-trained prior distributions with 2 training datasets (green) and 16 training datasets (blue) are shown along with the ground-truth prior distribution (red) for each GP parameter.}
    \label{fig:two-step-asymptotic-constant-sigvar-noisevar-synthetic}
\end{figure}

\subsection{Asymptotic behavior of the two-step pre-training method} Fig.~\ref{fig:two-step-asymptotic-constant-sigvar-noisevar-synthetic} shows on the left the estimated prior distribution parameter (mean $\mu$ for normal distribution and shape $\alpha$ for Gamma distribution) for GP parameters including length-scale, constant mean, signal variance and noise variance w.r.t. the number of training datasets. This is demonstrated only on the Synthetic Super-dataset since there is no known ground-truth prior distributions for HPO-B Super-dataset. 

In addition, the estimated prior distributions for each GP parameter with 2 training datasets and 16 training datasets along with the corresponding ground-truth prior distribution are shown on the right of Fig.~\ref{fig:two-step-asymptotic-constant-sigvar-noisevar-synthetic}. For most of the GP parameters (length-scale, constant mean and signal variance), the 
estimated prior distribution parameter shows decreasing variance and gets closer to the ground-truth value as the number of training datasets increases. 

The estimated prior distribution also gets much closer to the ground-truth prior distribution when 16 training datasets are used compared to only using 2 training datasets. For the noise variance parameter, the $\alpha$ parameter of Gamma distribution is estimated to be a different value than the ground-truth value, and the estimated prior distribution shows a different shape than the ground-truth prior distribution even when using 16 training datasets. Note that each dataset of the Synthetic Super-dataset contains only 10 sub-datasets. This small number of sub-datasets in each search space is likely part of the reason for the difference between the estimated noise variance prior and the ground-truth prior. The variance of the estimated $\alpha$ parameter of Gamma distribution for noise variance still decreases as the number of training datasets increases. The visualization shows that the estimated prior distributions for noise variance also put most of the probability density over values between [0, 0.0002] as the ground-truth prior distribution does despite the difference in their shapes.

\begin{figure}[h]
    \centering
    \includegraphics[width= \linewidth]{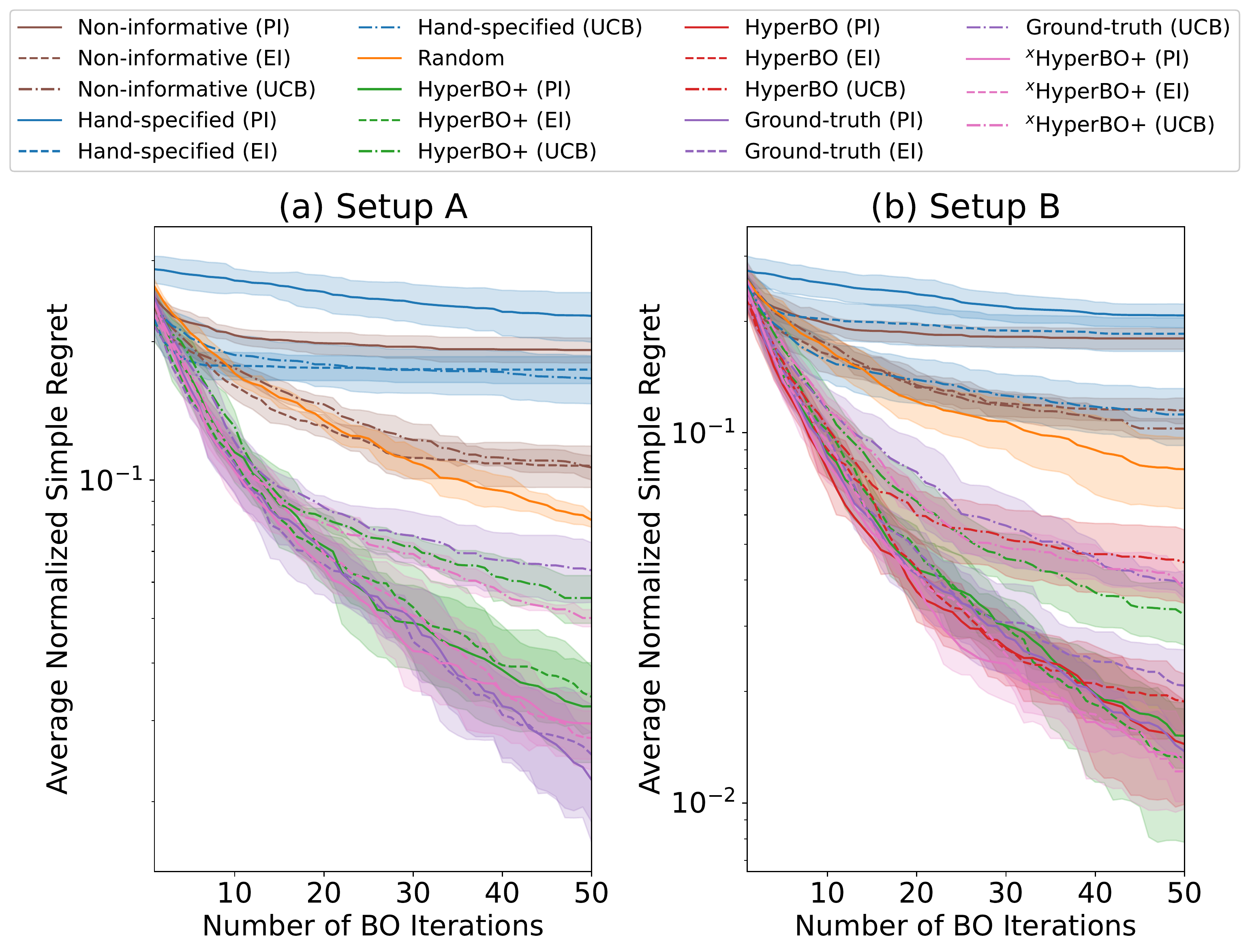}
    \caption{Average normalized simple regret of each method during BO in both Setup A and Setup B for the Synthetic Super-dataset. The averages are taken over 5 random seeds, and the highlighted areas show $\mathrm{mean} \pm \mathrm{std}$ for each method. Results for acquisition functions PI, EI, and UCB are shown here.}
    \label{fig:bo-regrets-all-acfun-synthetic}
\end{figure}

\begin{figure}[h]
    \centering
    \includegraphics[width= 0.9\linewidth]{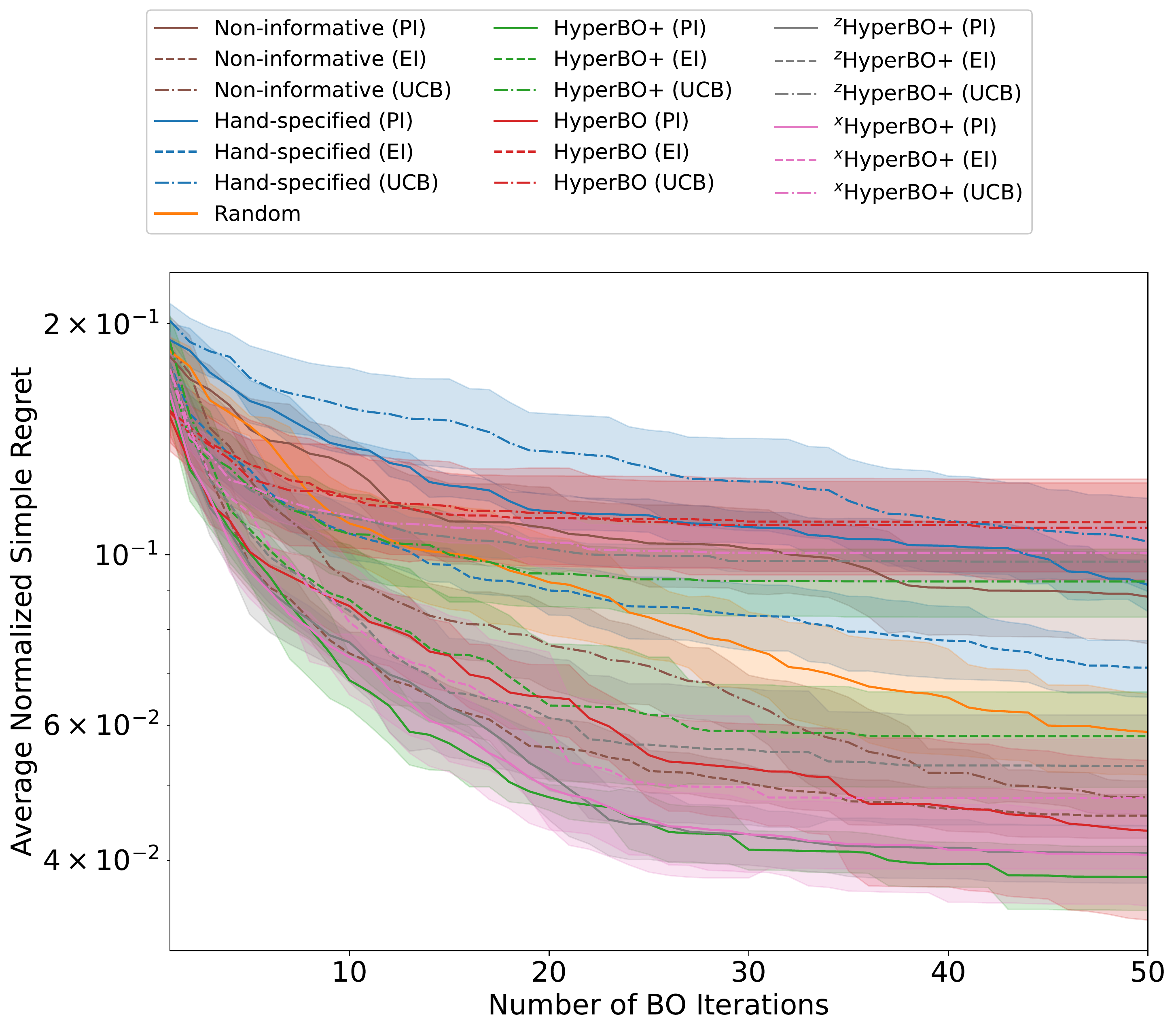}
    \caption{Average normalized simple regret of each method during BO for the HPO-B Super-dataset. The averages are taken over 5 random seeds, and the highlighted areas show $\mathrm{mean} \pm \mathrm{std}$ for each method. Results for acquisition functions PI, EI, and UCB are shown here.}
    \label{fig:bo-regrets-all-acfun-hpob}
\end{figure}

\subsection{BO performances with different acquisition functions} Fig.~\ref{fig:bo-regrets-all-acfun-synthetic} shows the average normalized simple regret of compared methods duing BO iterations with different acquisition functions in both Setup A and Setup B for the Synthetic Super-dataset and Fig.~\ref{fig:bo-regrets-all-acfun-hpob} shows the results with different acquisition functions in HPO-B Super-dataset. The acquisition functions included are \textit{Probability of Improvement} (PI), \textit{Expected Improvement} (EI), and GP-UCB. The PI acquisition function has parameter $\zeta = 0.1$, which means its target value is the maximum observation plus $\zeta$. The EI acquisition function uses the maximum observation as the target value. The GP-UCB acquisition function has parameter $\beta=3$. It can be observed that PI generally provides better BO performance for GP-based methods across the two super-datasets than EI and GP-UCB. This is in line with the finding by \cite{wang2017maxvalue} that PI can obtain high BO performance by setting good target values. Therefore we use PI as the acquisition function for BO results in the main paper. 

We can see that given the same acquisition function, HyperBO+ outperforms the non-informative hierarchical GP and the hand-specified hierarchical GP baselines and achieves a performance that is either similar to or better than that of HyperBO in most of the three cases (HPO-B Super-dataset and two setups for the Synthetic Super-dataset).

Notice that HyperBO+ aims to recover the ground-truth prior distribution, and as what we have shown here, we need a good acquisition function to unleash the best performance of BO given our pre-trained prior. Using misspecified priors, some sub-optimal acquisition functions may actually lead to better performance than an acquisition function that achieves better results with the ground-truth prior.

\end{document}